\documentclass{article}

\usepackage{microtype}
\usepackage{graphicx}
\usepackage{subfigure}
\usepackage{booktabs} 

\usepackage{hyperref}

\usepackage[accepted]{icml2024}

\usepackage{amsmath}
\usepackage{amssymb}
\usepackage{mathtools}
\usepackage{amsthm}

\usepackage[capitalize,noabbrev]{cleveref}

\theoremstyle{plain}

\theoremstyle{definition}

\theoremstyle{remark}

\usepackage[textsize=tiny]{todonotes}

\usepackage{algorithm}
\usepackage{algorithmic}
\usepackage[bb=boondox]{mathalfa}
\usepackage{listings}
\usepackage{multirow,tabularx}
\usepackage{amssymb}
\usepackage[T1]{fontenc}
\usepackage{wrapfig}
\usepackage{subcaption}
\usepackage{multirow}
\usepackage[normalem]{ulem}
\useunder{\uline}{\ul}{}

\newcommand{\framework}{OpenPAL} 
\newcommand{\Contra}{Contra}

\newcommand{\highlight}[1]{#1}

\icmltitlerunning{Building Open-Ended Embodied Agent via Language-Policy Bidirectional Adaptation}

\begin{document}

\twocolumn[
\icmltitle{Building Open-Ended Embodied Agent via Language-Policy \\ Bidirectional Adaptation}



\icmlsetsymbol{equal}{*}

\begin{icmlauthorlist}
\icmlauthor{Shaopeng Zhai}{equal,shlab}
\icmlauthor{Jie Wang}{equal,shlab}
\icmlauthor{Tianyi Zhang}{equal,shlab}
\icmlauthor{Fuxian Huang}{equal,shlab}\\
\icmlauthor{Qi Zhang}{equal,shlab}
\icmlauthor{Ming Zhou}{equal,shlab}
\icmlauthor{Jing Hou}{equal,tju,shlab}
\icmlauthor{Yu Qiao}{shlab}
\icmlauthor{Yu Liu}{shlab}
\end{icmlauthorlist}

\icmlaffiliation{shlab}{Shanghai AI Laboratory}
\icmlaffiliation{tju}{Tongji University}

\icmlcorrespondingauthor{Shaopeng Zhai}{zhaishaopeng@pjlab.org.cn}
\icmlcorrespondingauthor{Yu Liu}{liuyu@pjlab.org.cn}

\icmlkeywords{Machine Learning, ICML}

\vskip 0.3in
]



\printAffiliationsAndNotice{\icmlEqualContribution} 



\begin{abstract}
Building embodied agents on integrating Large Language Models (LLMs) and Reinforcement Learning (RL) have revolutionized human-AI interaction: researchers can now leverage language instructions to plan decision-making for open-ended tasks. However, existing research faces challenges in meeting the requirement of open-endedness. They typically either train LLM/RL models to adapt to a fixed counterpart, limiting exploration of novel skills and hindering the efficacy of human-AI interaction. To this end, we present OpenPAL, a co-training framework comprising two stages: (1) fine-tuning a pre-trained LLM to translate human instructions into goals for planning, and goal-conditioned training a policy for decision-making; (2) co-training to align the LLM and policy, achieving instruction open-endedness. We conducted experiments using Contra, an open-ended FPS game, demonstrating that an agent trained with OpenPAL not only comprehends arbitrary instructions but also exhibits efficient execution. These results suggest that OpenPAL holds the potential to construct open-ended embodied agents in practical scenarios.
\end{abstract}
\section{Introduction}
With the increasing prevalence of LLMs such as ChatGPT, researchers have progressively shifted their focus towards LLM-centered principles, building embodied agents that interact with humans to tackle open-ended tasks~\citep{khandelwal2022simple,huang2023voxposer}.
To achieve this target, we need to resolve the challenge of developing AI agents with the ability to continuously learn new skills, which is related to a domain commonly referred to as open-ended learning that is broadly categorized into two main factions: (1) pre-training LLMs to translate human-instructions into sub-tasks, for open-ended planning~\citep{wang2023voyager,ouyang2022training}, and (2) curriculum RL for open-ended control~\citep{team2021open,balduzzi2019open}.

For pre-trained LLMs, particularly those with closed source architectures, focus on resolving planning with general knowledge acquired during the pre-training stage~\citep{wang2023voyager}.
However, they share shortcomings like relying on task-oriented and hand-crafted prompting, struggling to comprehend interactions in special contexts such as games and be incompetent for high real-time requirements due to inefficient model computation.
In contrast, curriculum RL conducts open-ended learning in an end-to-end manner, developing in diverse methodologies such as population-based RL~\citep{team2021open}, goal-conditioned RL (GCRL)~\citep{liu2022goal} and etc.
Despite RL excels in learning novel skills compared to rule-based control, it lacks the direct capability for interaction with humans.
\begin{figure*}[ht]
	\centering	
        \includegraphics[width=.9\textwidth]{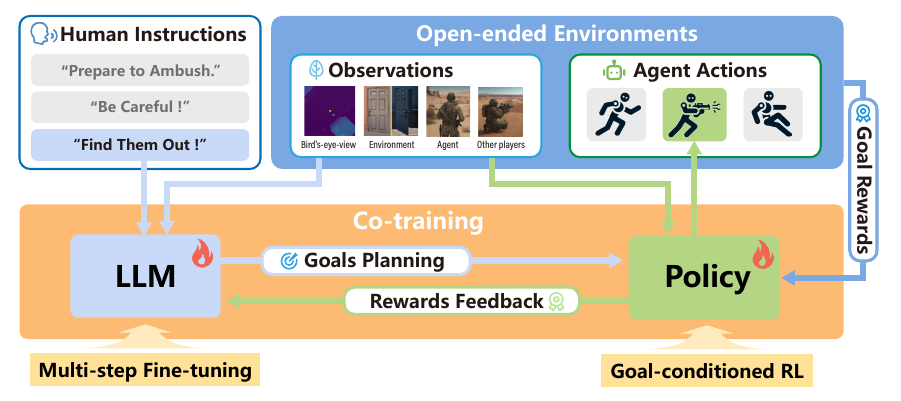}
	\centering
	\caption{Overview of co-training in \framework{}. The Policy and LLM is pre-trained with multi-step fine-tuning and goal-conditioned RL, respectively. Then, the co-training aligns them towards achieving instruction open-endedness.}
	\label{fig:co_training}
\end{figure*} 
To leverage advantages from both sides, i.e., being capable of interacting human and AI in solving real-time tasks towards open-endedness, an ideal implementation is to integrating LLM for planning and RL policy for decision making.
However, existing studies in this domain have focused on improving training efficiency or reducing interaction costs by either independently training the RL policy or LLM~\citep{hu2023enabling,du2023guiding} to adapt the other, resulting in overfitting and failing to explore novel skills in specific environments that necessitate specialized knowledge and falling short of achieving true open-endedness. Despite previous work resolve this issue with LLM-based re-planning~\citep{wang2023voyager,wang2023describe}, it is inefficient for high-dimensional tasks and the re-planning is still in the existing range of strength.

To address the above challenge, we propose a co-training framework, \framework{}, strunctured as a two-stage learning process to implemente bi-directional adaptation. This design enables the RL policy continuously explore novel skills whiling align the LLM and the policy towards achieving instruction open-endedness. In the first stage, we separately train a ChatGLM-6B~\citep{du2022glm} as a planner (or goal generator) $G_{llm}$ and policy $\pi_g$, where $G_{llm}$ generates goals with given instructions and environment context, and $\pi_g$ learns to execute goals. To achieve that, we propose multi-step fine-tuning a pre-trained LLM with GPT-4-generated instructions and goals, and open-ended goal generation to learn a goal-conditioned policy. In the second stage, we implement co-training to align $G_{llm}$ (planning) and $\pi_g$ (decision-making), as illustrated in \Cref{fig:co_training}. This aims to achieve instruction open-endedness, aligning the instruction space with the open-ended goal space that the agent achieved. Specifically, we implement the co-training as an interleaved execution of (1) Reinforcement Learning with Agent Feedback (RLAF) for $G_{llm}$ and (2) GCRL for $\pi_g$ with goals generated by $G_{llm}$, where RLAF centers around rewarding $G_{llm}$ with agent feedback and goal execution.
This two-staged approach optimizes the LLM for comprehending environment context under the consideration of decision-making, while concurrently enhancing decision-making for goals aligned with human instructions. 
For evaluation, we employ \Contra{}, an open-ended FPS game. The results demonstrate that \framework{} achieves a high goal completion ratio for open-ended human-AI interaction.
\section{Background}
\paragraph{Goal-conditioned Reinforcement Learning.}\label{sec:goal_conditioned_rl}
Formally, GCRL could be formulated as a goal-augmented Markov Decision Process $\mathcal{M}$~\citep{liu2022goal}.
Denoting $\mathcal{M}$ a tuple $\langle \mathcal{S}, \mathcal{A}, \mathcal{G}, \mathcal{P}, \mathcal{R}, \gamma \rangle$, where $\mathcal{S}$, $\mathcal{A}$, $\mathcal{G}$ the state, action and goal spaces, respectively.
In general, $\mathcal{G}$ is a projection of $\mathcal{S}$, i.e., $\mathcal{G} = \textsc{Proj}(\mathcal{S})$.
$\mathcal{P}$ defines the state transition probabilities, i.e., $\mathcal{P}: \mathcal{S} \times \mathcal{A} \rightarrow \Delta(\mathcal{S})$, where $\Delta(\cdot)$ a distribution.
$\mathcal{R}: \mathcal{S} \times \mathcal{A} \times \mathcal{G} \rightarrow \mathbb{R}$ defines the reward function $r(s,a,g)$.
At the beginning of an trajectory $\tau$, a goal $g$ is sampled from a distribution $P_g$, which generally defines a task for the agent to execute. As for decision making, $\pi$ denotes a policy as $\pi: \mathcal{S} \times \mathcal{G} \rightarrow \Delta(\mathcal{A})$, which is a distribution over the action space. 
To solve a $\mathcal{M}$, or achieve open-endedness in other words, an agent with policy $\pi$ needs to maximize its accumulative reward over the goal space as $\mathbb{E}_{a_t \sim \pi(\cdot | s_t, g), (s_t,a_t) \sim \tau, g \sim P_g}\left[ \sum^T_{t=0} \gamma^t r(s_t,a_t,g) \right]$,
where $\gamma \in [0, 1)$ discounts a reward at each time step to ensure the convergence.
Normally, $r(s_t,a_t,g)$ is binary as
\begin{equation}\label{eq:goal_reward}
    r(s_t,a_t,g) = 
    \begin{cases}
        1 & \textsc{Proj}(s_{t+1})=g\\
        0 & \text{otherwise}
    \end{cases}.
\end{equation}
To approximate $\sum^T_{t=0} \gamma^t r(s_t,a_t,g)$, GRL suggests using the Universal Value Function Approximator (UVFA) $V(s,g)$. As for the solving of open-endedness, there are critical challenges that need to be resolved, including but not limited to:
(1) $P_g$ is agnostic to an agent, and (2) $\mathcal{G}$ is uncountable and continuous.
To resolve these issues, existing research is centered on curriculum RL~\citep{weng2020curriculum}, i.e., automatically discovering novel goals from past learning. Hindsight goal relabelling~\cite{andrychowicz2017hindsight,fang2019curriculum,zheng2022online,schaul2015prioritized} implicitly implements curriculum learning by introducing a prioritized replay mechanism and performs high data efficiency.
Despite numerous curriculum approaches, the sample inefficiency due to the setting of binary reward signals (\Cref{eq:goal_reward}) hinders policy learning.
As a solution, existing research leverages reward shaping~\citep{ng1999policy,ecoffet2021first,ding2023magnetic,trott2019keeping} which is a straightforward and efficient idea.

\paragraph{LLMs and Human-AI Interaction.}
LLMs~\citep{brown2020language,openai2023gpt4} a class of neural networks that execute in auto-regressive for text generation.
Given a sequence of text tokens with length $t$ as $x_{1:t} = (x_1,\dots,x_t)$, the generation of a next token $x_{t+1}$ could be formulated as sampling from a probabilistic model $P(\cdot \vert x_{1:t})$.
As for the training of LLMs, the target is equivalently to find a parameter set $\theta_{\text{LLM}}$ which satisfies the optimal generation, i.e., $\theta_{\text{LLM}} = \arg\max_{\theta} \mathbb{E}_{x_{1:t},x_{t+1}}P(x_{t+1} \vert x_{1:t};\theta)$.
Beyond the research of LLMs, it is attractive to leverage LLMs as an interface to interact human with agents~\citep{jiang2023vima,hermann2017grounded}.
We can roughly reformulate the generation as $x_{t+1}\sim P(\cdot \vert \iota, x_{1:t})$ for human-AI interaction, where $\iota$ the language instruction as a prompt, $x_{1:t}$ the context.
For the cases have low real-time requirements, $x_{t+1}$ is a control signal for decision making~\citep{brohan2022rt,zitkovich2023rt,nakano2021webgpt}.
While for the cases have high real-time requirements, $x_{t+1}$ is a goal and will be fed to a controller to guide the decision making~\citep{wang2023describe,wang2023voyager}.
Our study falls within the latter situation, building open-endeded embodied agents in the cases with high real-time requirements.

\section{The \Contra{}: A Battle Royale FPS Game}
\Contra{} seamlessly merges the last-man-standing gameplay dynamics with the survival, exploration, and scavenging elements inherent in first-person shooting games~\citep{gautam2021battle}.
It unfolds with multiple hostile teams, necessitating players to collaborate with teammates, withstand adversaries, and strive to outlast others in the ever-changing arena. The agent's objectives encompass individual survival and the elimination of encountered enemies. An agent in \Contra{} mandates a sequential acquisition of skills, starting from fundamental abilities like running and item collection. As the learning proceeds, an agent must master more intricate skills such as evading enemy projectiles and coordinating tactics with teammates, an open-ended learning process.
\highlight{The primary rationale behind choosing \Contra{} as our testbed lies in its possession of proprietary knowledge not encompassed in general knowledge repositories. Consequently, we employ Reinforcement Learning (RL) for knowledge discovery, and co-training to align the Language Model (LLM) and RL in comprehending the environment.}
\section{A Co-training Framework: \framework{}}
Considering the training in the context of open-ended learning is extensive data-thirsty, we first introduce two critical engineering designs to enhance training efficiency. Specifically, \framework{} incorporates a distributed RL framework inspired by AlphaStar~\citep{vinyals2019grandmaster} with modifications, resulting in the formation of the \emph{Actor-League-Learner} architecture. In this architecture, the \emph{League} is responsible for distributing rollout tasks to a cluster of \emph{Actors} (CPU nodes) for data collection and evaluation, while optimization tasks are delegated to the \emph{Learner} (GPU node) for policy updates. This distributed approach significantly enhances rollout throughput, thereby improving overall training efficiency.
Another efficiency challenge stems from the iterative development of \Contra{}.
During the period of our research, \Contra{}'s environmental attributes continuously change as the programming development. Thus, policy retraining would be necessary if there is no explicit intervention.
To reduce such an extra computation burden, we employ surgery~\cite{openai2019dota} to retain learned skills at the lowest training cost, enabling adaptation to a changing observation/goal space while ensuring compatibility with network inputs. Detailed information on the distributed RL framework can be found in \Cref{app:training_system}, and version changes are listed in \Cref{tab:surgery_events}. In the following content, we will introduce \framework{} in two stages, including the independent training at stage I (\Cref{sec:mastering_novel_skills} $\sim$ \ref{sec:pretrain_llm}) and the co-training at stage II (\Cref{sec:co_training}).

\subsection{Exploring Basic Skills via Non-goal RL}\label{sec:mastering_novel_skills}
In the realm of GCRL, the prevalent approach involves curriculum learning a goal-conditioned policy from scratch, learning goal execution while exploring goals.
However, it maybe inefficient for an agent to explore the whole goal space when there is a lack of prior knowledge of the goal space.
Thus, we opt for leveraging non-goal RL for basic skill learning before goal-conditioned learning.
For the implementation, we employ Proximal Policy Optimization (PPO)~\citep{schulman2017proximal} with fine-grained reward shaping as
\begin{equation}\label{eq:non_goal_reward_function}
    r(s,a) = \lambda_1 r^{b}(s,a) + \lambda_2 r^{oa}(s,a),
\end{equation}
where $r^{b}$ for skill learning which focuses on targeting the agent towards wining and surviving as long as possible, is a linear combination of diverse behavior factors, $r^{oa}$ encourages the agent to avoid obstacles due to the agent is not sensitive to obstacles when navigating, and $\lambda_1$, $\lambda_2$ the factors to weight the contribution of each item. The details of the reward construction are included in \Cref{tab:reward_weight}, \Cref{app:env}.
Then, the value function for estimating $\sum^T_{l=t}\gamma^{l-t} r(s_l,a_l)$ is implemented as a multi-head network and shares the backbone of policy, i.e., $V(s_t) = \lambda_1 V^{b}(s_t) + \lambda_2 V^{oa}(s_t)$, where $V^{b}(s_t)$ and $V^{oa}(s_t)$ approximate $\sum^T_{l=t}\gamma^{l-t}r^{b}(s_l,a_l)$ and $\sum^T_{l=t}\gamma^{l-t} r^{oa}(s_l,a_l)$, respectively.

\subsection{Learning a Goal-conditioned Policy}
\label{sec:stage_cgrl}
We construct the goal space using various state attributes which can be determined and effected by interaction. In detail, they are (1) \textbf{agent private states} that can be directly changed by the agent or other players, such as firing, walking, etc. (2) \textbf{enemies states} that can be effected through the agent interactions, such as knock down an enemies; and (3) \textbf{teammates states} that can be effected by the interaction between the agent and its teammates.
We summarize them in \Cref{tab:features_summary}.
With the above consideration, we further model each attribute as a sub-goal space $\mathcal{G}^i$ with multiple candidates that can be expressed as a set of normalized indices $\{ \frac{j}{|\mathcal{G}^i|} \vert j=0,\dots,|\mathcal{G}^i|\}$, where 0 an invalid attribute value indicates the corresponding attribute not be selected. For the goal space, there are 68 sub-goal spaces that shape it as $\mathcal{G} = \Pi^{68}_{i=1}\mathcal{G}^{i}$.
Obviously, $g$ which comprises of more valid sub-goals, the more difficult to complete for the policy.

\paragraph{Open-ended Goal Generation.}
Among existing GCRL research, hindsight goal relabelling and generation~\citep{andrychowicz2017hindsight,ren2019exploration} are effective goal-conditioned learning methods that advantage from a free of goal prior, compared to explicit curriculum.
However, there is a limitation of in-distribution goal exploration~\citep{bai2019guided}, i.e., policy learning and goal exploration shares the same training dataset, which is inefficient in exploration as the range of goals are limited by the scale of samples.
Comparatively, if we can model the goal distribution, we can not only achieve data efficiency akin to that of hindsight goal generation, but also progressively attain an open-ended goal space by adjusting the goal distribution.
Therefore, we train a neural-based goal generator $G_{op}$ over a dataset of trajectory segments $\mathcal{D}_{\tau} =\{\tau\}$ explored by the well-trained policy $\pi^{\star}$ from \Cref{sec:mastering_novel_skills} as it is trained for exploring basic skills.
We assume that a goal $g$ corresponding to a given initial state $s$ can be represented by a 3-tuple $x = (s, \Delta t, \Delta V)$, where $\Delta t$ the time slot required to achieve $g$ starting from $s$, and $\Delta V$ a vector of state values from $s$ to $g$ with the consideration of representation.
As a result, we train the goal generator $G_{op}$ to take input in the form of $x$, thereby allowing variations in $\Delta t$ and $\Delta V$ to yield different goals for a given state $s$.
For an implementation, we firstly construct a dataset $\mathcal{D}_{x,g} = \{\left(x, g\right)\}$ from $\mathcal{D}_{\tau}$, where each item in $\mathcal{D}_{x,g}$ satisfies:
\begin{align}
    s &\sim \textsc{Uniform}(\tau_{:150}),\\\nonumber
    g &= \textsc{Proj}(s'),\quad s'\sim P(s')=\frac{V^{b}(s' \vert s' \in \tau_{-20:})}{\sum_{s' \in \tau_{-20:}} V^{b}(s')},\\\nonumber
    \Delta t &=\text{ the time slot from }s\text{ to }s',\\\nonumber
    \Delta V &= \left[V^{b}(s), V^{b}(s')\right].
\end{align}
$\tau_{:150}$ the first 150 states of $\tau$, $\tau_{-20:}$ the last 20 states.
Then we train $G_{op}$ with a MSE loss as $\min_{G_{op}} \mathbb{E}_{\mathcal{D}_{x,g}} \left[\lVert G_{op}(x) - g\rVert_2\right]$.
While varying $\Delta t$ and $\Delta V$ produces diverse goals, it remains challenging to comprehensively cover the entire goal space corresponding to a given state. As a supplement, we propose integrating goal generation with a uniform sampler, denoted as $G_{rnd}$, which randomly samples goals from the goal space $\mathcal{G}$ using $G_{rnd} = \Pi^{68}_{i=1}\textsc{Uniform}(\mathcal{G}^i)$. This results in a goal generation $g \sim G_{rnd} \cup G_{op}$.

\paragraph{Intrinsic Reward Shaping.}
As introduced in the aforementioned, a critical challenge hinders the goal completion is sparse rewarding.
To mitigate this, we extend the reward function in \Cref{eq:non_goal_reward_function} with an intrinsic reward $r^{g}(s_t,a_t,g)$ that evaluates the degree of goal completion. \Cref{eq:goal_reward} shows the calculation of $r^{g}(s_t,a_t,g)$ as the Euclidean norm difference between two consecutive states and a goal as
\begin{equation}\label{eq:goal_reward}
    r^{g}(s_t,a_t,g)= \lVert g - \textsc{Proj}(s_{t-1}) \rVert_p - \lVert g - \textsc{Proj}(s_t) \rVert_p,
\end{equation}
where $\Vert \cdot \Vert_p$ indicates the $p$-norm. This reward provides a denser reward signal at each time step to the agent about its proximity to the goal, offering more nuanced information than a binary signal indicating whether it has reached the goal or not. In our current implementation, we set $p=1$.
Thus, the reward function for GCRL is formulated as
\begin{equation}\label{eq:goal_learning_reward}
    r(s,a,g) = r(s,a) + \lambda_3 r^{g}(s,a,g),
\end{equation}
where $r(s,a)$ comes from \Cref{eq:non_goal_reward_function}.
And for the value function corresponds to \Cref{eq:goal_learning_reward}, we extend the multi-head $V(s_t)$ with a new value head $V^g(s_t,g)$ as $V(s_t,g) = V(s) + \lambda_3 V^g(s_t,g)$, where $V^g(s_t,g)$ approximates $\sum^T_{i=t}\gamma^{i-t}r^{g}(s_t,a_t,g)$.

\paragraph{Avoiding Policy Degeneration.}
Let $\pi^{\star}$ denote the well-trained policy from the non-goal reinforcement learning step. However, we have observed a performance degeneration on basic skill execution when continuing the training of the goal-conditioned policy $\pi_{g,\theta}$ starting from $\pi^{\star}$. 
This is attributed to two aspects: (1) catastrophic forgetting on the basic skills as the goal-conditioned learning continues; (2) a change in the input of the policy network from $s$ to $(s,g)$, where $g$ introduces interference in decision-making, as the policy has not encountered goal inputs during non-goal-conditioned training.
To address thes issues, we propose a modification to the goal-conditioned policy learning objective by introducing a KL-divergence regularizer, and introduce 20\% workers for non-goal policy learning to avoid catastrophic forgetting. This regularizer quantifies the distance between $\pi^{\star}$ and $\pi_{g,\theta}$ when $\pi_{g,\theta}$ conditioned on $g=\emptyset$ as it is equivalently to non-goal policy:
\begin{equation}
\label{eq:kl_policy}
\max_{\theta} \mathbb{E}_{(s,g)}\left[J(\pi_{g,\theta}) -\mathbb{1}_{g=\emptyset} \cdot D_{KL}(\pi^{\star} \parallel \pi_{g,\theta})\right].
\end{equation}
$J(\pi_{g,\theta})$ the policy loss in PPO, and $\mathbb{1}_{g=\emptyset}$ indicates that the KL-divergence term is only activated when an empty goal input for $\pi_{g,\theta}$.
\Cref{alg:rl}, \Cref{appendix_algorithm} summarizes the learning process.
Furthermore, we observed that occasionally sampling experience from $\pi^{\star}$ to train $\pi_{g,\theta}$ can also relieve the degeneration.

\subsection{Fine-tuning a LLM-based Goal Generator}
\label{sec:pretrain_llm}
Let $\mathcal{I}$ represent the set of natural language instructions, and $\mathcal{O}$ the set of abstracted environment states in text. Our objective is to fine-tune a pre-trained LLM as a goal generator, denoted as $G_{llm}: \mathcal{O} \times \mathcal{I} \rightarrow \mathcal{G}$, which means $G_{llm}$ generates a goal relevant to a given instruction with the consideration of current environment context, i.e., a state abstraction.

\paragraph{Dataset Construction.}
To achieve that, we first construct $\mathcal{O}$ using states collected by the $\pi_{g,\theta}$. Each abstraction $o \in \mathcal{O}$ encapsulates essential state features of its corresponding state $s$, and the extraction rules are outlined in \Cref{appendix:environment_state_abstraction}.
For the creation of $\mathcal{I}$, we leverage various instruction generation to ensure its diversity and scalability, aligning with our overarching goal of achieving open-endedness in the instruction space. Specifically, $\mathcal{I}$ is derived from four types. 
Most of these are formulated through a tuple of an initial state and a target state/trajectory collected by $\pi_g$, which aims to align $G_{llm}$ and $\pi_g$ at environmental comprehension. Then, we leverage this data and GPT-4~\citep{openai2023gpt4} to generate appropriate instruction. This instruction aims to direct from the specified initial state to the intended target state, and CoT~\citep{wei2023chainofthought} is deployed to enhance performance.
Specifically, the four types of instruction generation are
(1) \textbf{$\mathcal{I}_H$ (Human Instructions, HI)}: human-annotated instructions;
(2) \textbf{$\mathcal{I}_S$ (State Instructions, SI)}: GPT-4-generated instructions by giving a tuple of states $(s,s')$ where the $s$ the initial state that sampled from agent trajectories and $s'$ the target state that is manually constructed by modifying features of the $s$;
(3) \textbf{$\mathcal{I}_A$ (Agent Instructions, AI)}: GPT-4-generated instructions by giving a pair of $(s,\tau)$ where $s$ the initial state, $\tau$ the agent trajectory;
and (4) \textbf{$\mathcal{I}_R$ (Random Instructions, RI)}: a mixture of the above three instruction sets to form a supplementary dataset.
By accompanying $\mathcal{O}$ with $\mathcal{I}$, we further construct $\mathcal{D}_x= \left\{(o, \iota) \vert (o,\iota) \in \mathcal{O} \times \mathcal{I}\right\}$. Subsequently, we employ GPT-4 to generate appropriate goals $\hat{\mathcal{G}}$ using $\mathcal{D}_x$ as labeled data for training $G_{llm}$, resulting in a dataset $\mathcal{D} = \{(o,\iota,g) \vert (o,\iota,g) \in \mathcal{O} \times \mathcal{I} \times \hat{\mathcal{G}}\}$.
To ensure that the goals generated by GPT-4 conform to the format we want, a comprehensive prompt engineering endeavor was conducted to establish a set of predetermined rules for GPT-4.
The rule-based prompts that guide GPT-4's responses are documented in \Cref{tab_rule}, with examples of prompts for generation provided in \Cref{tab_prompt}.
\paragraph{Multi-step Fine-tuning.}
We fine-tune ChatGLM-6B with LoRA~\citep{hu2021lora} in three steps, as illustrated in \Cref{fig:LLM_RL_framework}. The steps include 
(1) \textbf{CoT-assisted fine-tuning (CoFT)}: we split the CoT steps of building $\mathcal{I}$ into independent training data, aiming to expand the volume of training data as well as enhance the goal generator's reasoning and understanding to $\mathcal{D}_x$;
(2) \textbf{Supervised Fine-tuning (SFT)}: strictly formatting the LLM-generated goals and further improving the accuracy;
and (3) \textbf{Ensemble Fine-tuning (EFT)}: multiple checkpoints of $G_{llm}$ are utilized to generate goal candidates for each $(o,\iota) \in \mathcal{D}_x$, then sub-goals with highest counts are reconstructed as a ground goal to fine-tune the model to enhance the generation.

\subsection{Collaborative Training}\label{sec:co_training}
After completing the above training steps, we obtained a well-trained goal generator $G_{llm}$ and goal-conditioned policy $\pi_{g}$ that satisfactorily adhere to their respective goal distributions.
However, an inconsistency persists between $G_{llm}$ and $\pi_{g}$ stemming from their independent training objectives, where $G_{llm}$ aimed to generate goals that satisfy given instructions, and $\pi_{g}$ focused on exploring goals.
Therefore, we introduce co-training to address the aforementioned issue ensuring that the goals generated by $G_{llm}$ are not only linguistically sound but also aligned with the capabilities of $\pi_g$. We formulate the co-training as follows:
\begin{align}\label{eq:co_training_objective}
  \begin{cases}
      \pi_g &= \arg\max_{\pi_g}\mathbb{E}_{g \sim P_{\mathcal{G}\vert G_{llm}}}\left[ V_{\pi_g}(s,g) \right]\\
      P_{\mathcal{G} \vert G_{llm}} &= \arg\max_{P_{\mathcal{G} \vert G_{llm}}}\mathbb{E}_{g \sim P_{\mathcal{G} \vert G_{llm}}}\left[ V_{G_{llm}}(s,g) \right],\\
  \end{cases}
\end{align}   
where $P_{\mathcal{G} \vert G_{llm}}$ the goal distribution conditioned by $G_{llm}$,
$V(s,g)$ denotes an approximate evaluation for $\pi_g$ or $G_{llm}$, in general, a state value function.
It is noteworthy that our co-training framework is close to a hierarchical reinforcement learning framework (HRL)~\citep{vezhnevets2017feudal}, where the Manager (comparable to $G_{llm}$) plans goals for the learning of the Worker (comparable to $\pi_g$), with RL being performed for each.
Inspired by HRL, we implement co-training by integrating the goal-conditioned training of $\pi_g$ and Reinforcement Learning with Agent Feedback (RLAF) for $G_{llm}$.
RLAF is built upon PPO, with a reward shaping that considers
(1) $R^f$ the evaluation of goal completion, where a high reward indicates that a goal is completed or the reachable probability from current state;
(2) $R^e$ the evaluation of crucial sub-goal completion, which involves examining cases by pairing instructions in a batch with a set of essential sub-goals;
(3) $R^m$ the evaluation of outputting the proper goal format, with the LLM being penalized based on edit distance.
Then, we can express the reward function as $R(s,\iota,g) = R^f(s,g) + R^e(s,g) + R^m(s,g)$ and \Cref{app:cot_reward} includes more details.
We observed the training will lead $G_{llm}$ and $\pi_g$ compromise to a local optimal, i.e., $G_{llm}$ comforts a high completion ratio for $\pi_g$ but neglect consistency with instructions, and $\pi_g$ simultaneously rewards $G_{llm}$ with a high completion ratio.
Furthermore, as the policy training continuing, the evaluation for goal generation is out-date.
To fix this issue, we propose a periodic reset for the RLAF, i.e., the parameters of the $G_{llm}$ will be reset for every set number of steps to avoid being trapped in a local convergence, achieving enhanced goal completion, and keeping goals consistent with human instructions.
Considering the training efficiency, we conduct LoRA~\citep{hu2021lora} to update the model weights for $G_{llm}$. \Cref{fig:LLM_RL_framework} illustrates the whole training process, and \Cref{alg:co_train} summarizes the corresponding pseudo-code.
\begin{algorithm*}[h]
	\caption{\textsc{Collaborative Training}}
	\label{alg:co_train}
	\begin{algorithmic}[1]
        \STATE  \textbf{Input:} $\theta$ the parameters of $\pi_{g,\theta}$; $\beta=\{\beta_{llm},\beta_{LoRA}\}$ for $G_{llm,\beta}$, $\beta_{llm}$ the pre-trained parameters of $G_{llm}$; $\beta_{LoRA}$ the fine-tuned LoRA parameters of $G_{llm}$; $\mathcal{I} = \mathcal{I}_{H} \cup \mathcal{I}_{S} \cup \mathcal{I}_{A} \cup \mathcal{I}_R$ the instruction set

        \STATE Reloading $\theta$, $\beta_{llm}$, merging $\beta_{LoRA}$ into $\beta_{llm}$
        \FOR{loop=1, 2, ... }{
            \STATE Initialize a new $\beta_{LoRA}$ and $\beta = \{\beta_{llm}, \beta_{LoRA}\}$
            \FOR{iteration=1, 2, ..., n}{
            \STATE Agents from a batch of workers send states $\{s_j \vert j=1,\dots,m\}$ to $G_{llm}$
            \STATE Random sample a batch of instructions: $\mathcal{I}_{train} = \{\iota_j \vert j=1,\dots,m\} \subset \mathcal{I}$}
            \STATE Generate goals in string with LLM: $\mathcal{G}_s=\{g_j \sim G_{llm,\beta}(s_j,\iota_j) \vert j=1,\dots,m\}$ and parse $\mathcal{G}_s$ to formatted goals: $\mathcal{G}$ 
            \STATE Distribute $\mathcal{G}$ to agents $\pi_{g,\theta}$, then collect trajectories $\{\tau_j\}$ and returns $\{R_j\}$ to form
            \begin{equation*}
                (\tau,R)=\{(\tau_j,R_j) \vert j=1,\dots,m\},\text{where }\tau_j=\{g_j,s_1,a_1,r_1,\dots,s_{T_j},a_{T_j},r_{T_j}\},R_j=\sum^{T_j}_{t=1}r_t
            \end{equation*}
            \STATE Update $\theta$ with \Cref{eq:kl_policy} and $(\tau, R)$
            \STATE Filter completed goals $\mathcal{G}_{c}$ from $\mathcal{G}$, extract $R$ the set from $(\tau, R)$ as agent feedback rewards $R^f$
            \STATE Compute examination reward by evaluating crucial sub-goal completion as $R^e=\textsc{Reward}_e(\mathcal{G}_{c},\mathcal{I}_{S} \cap \mathcal{I}_{train},\mathcal{G})$, and formatting reward: $R^m=\textsc{Reward}_m(g_s,g)$, refer to \Cref{app:cot_reward}
            \STATE Update $\beta_{LoRA}$ with PPO, and merge updated $\beta_{LoRA}$ into $\beta_{llm}$
            \ENDFOR
        }
        \ENDFOR
	\end{algorithmic}
\end{algorithm*}
\section{Experiment}
We conduct empirical experiments to evaluate the efficacy of both stages of our proposed \framework{}. To make the \Contra{} satisfy the learning requirements, we give well-designed
spaces and reward functions as follows.

\paragraph{Observation Space.}
The observation space encompasses many factors, such as unit features detailing the agent states, those of other players and environmental features capturing interaction events. Additionally, an agent-centric RGB bird's-eye-view (BEV) of the local environment is considered.
\Cref{tab:features_summary} includes detailed information.

\paragraph{Action Space.}
The action space is implemented on top of \Contra{}'s micro-operation API, comprising a collection of multi-grained actions. These actions range from fine-grained movements, such as six-degrees-of-freedom movement and weapon usage, to compound actions in coarse-grained categories, such as firing at a target, and each action is executed over a duration of 200ms, hence the control frequency is 5Hz. The total size of the action space is 54. Further details in \Cref{tab:action_space}.

\paragraph{Reward Functions.}
A comprehensive representation is employed for the reward function, considering various factors contributing to goal-conditioned policy learning. These factors are organized as a linear combination to formulate the reward function. Furthermore, we determine the weights for the combination with a two-fold principle: (1) assigning weights to reward items based on their scales and emphasizing important factors; (2) dynamically adjusting weights in response to learning feedback, such as decreasing or increasing the weights of corresponding factors. Additional information is available in \Cref{app:env}.

\subsection{Evaluating Goal-conditioned RL}
\begin{figure*}[th]
\centering
\subfigure[]{
\includegraphics[width=0.65\columnwidth]{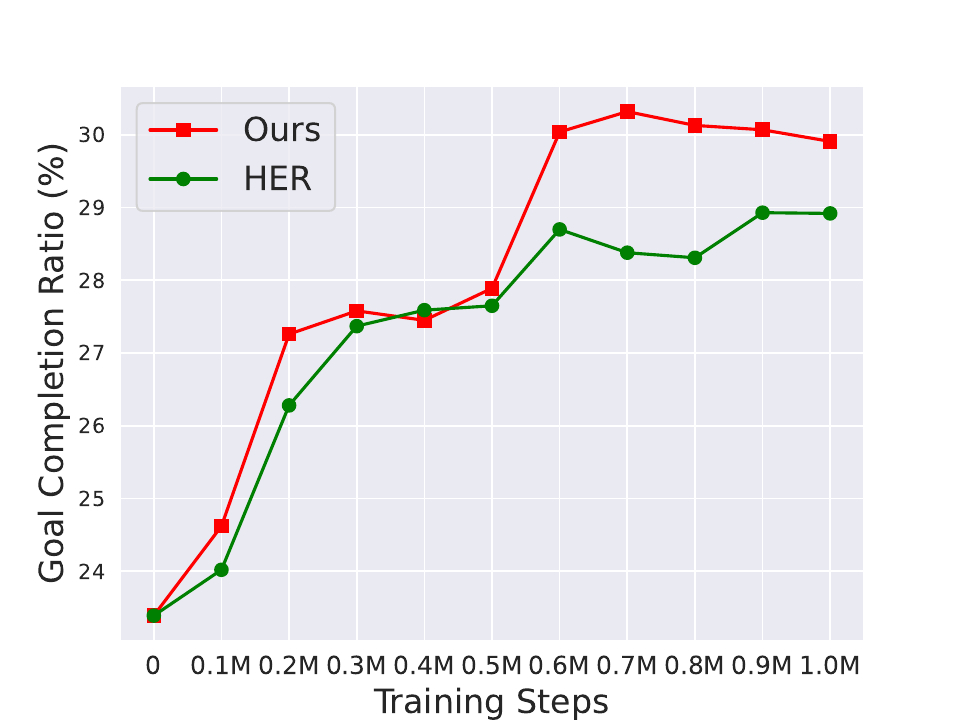}
\label{fig:rl_pretrain_goal_train}
}
\subfigure[]{
\includegraphics[width=0.65\columnwidth]{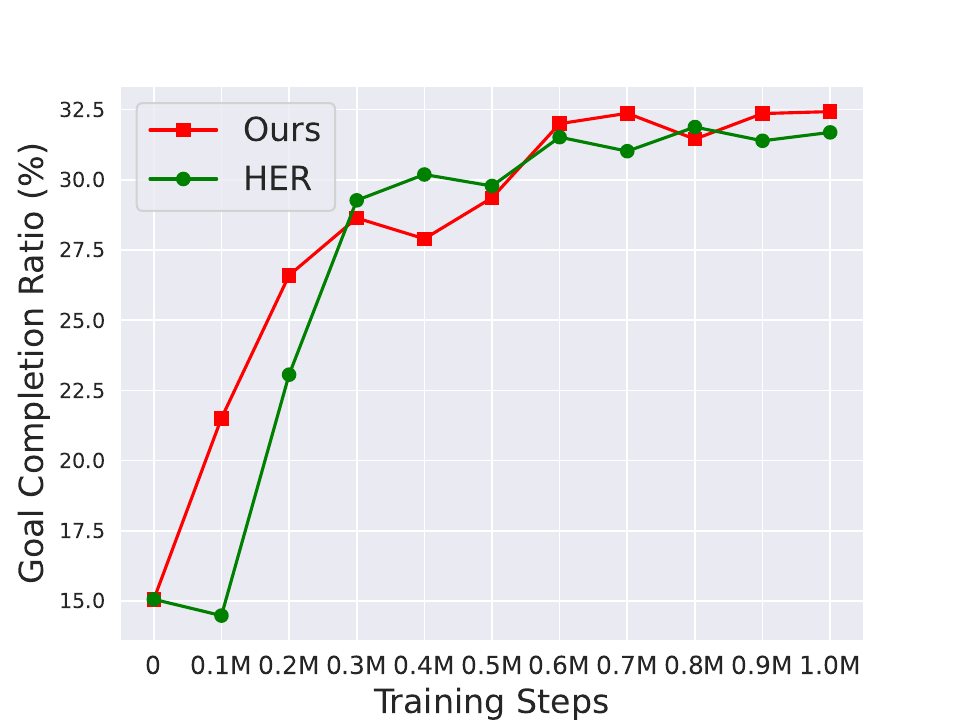}
\label{fig:rl_pretrain_goal_test}
}
\subfigure[]{
\includegraphics[width=0.65\columnwidth]{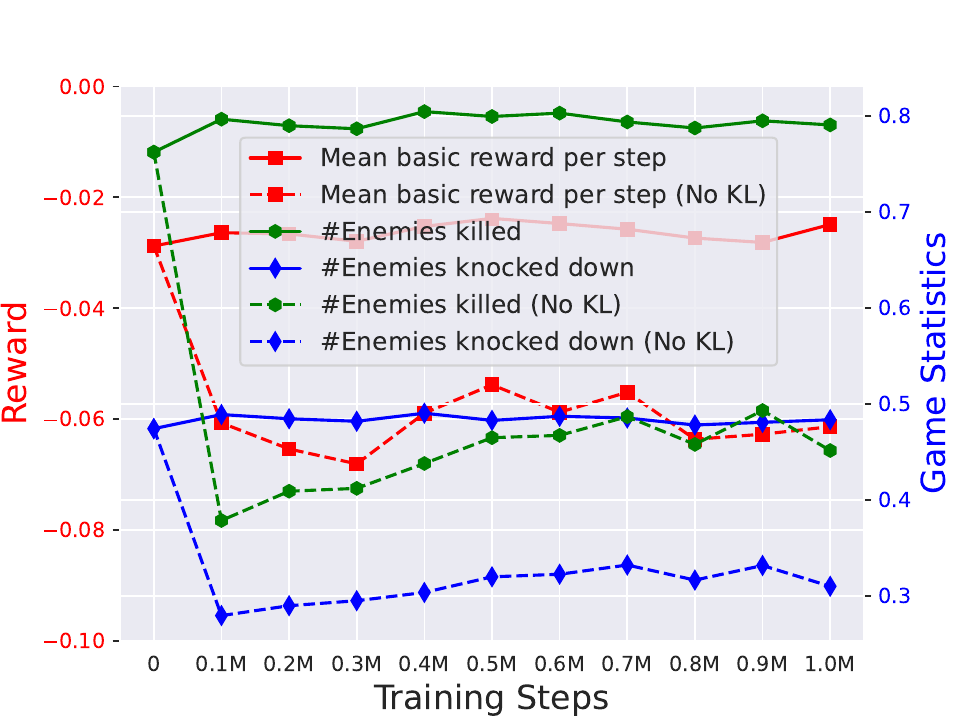}
\label{fig:degradation_reward_kil}
}
\caption{(a) The goal completion rate on training dataset;  (b) The goal completion rate on unseen goals, i.e., the test dataset; (c) The evaluation of policy learning in cases of w/ and w/o KL-divergence regularizer.}
\end{figure*}

We evaluate the $\pi_g$ of stage I from three distinct perspectives to verify the open-endedness achieved on $\mathcal{G}$: (1) the completion ratio, (2) generalization capability concerning unseen goals, and (3) robustness when integrating goal-conditioned learning atop non-goal learning.
Given that GCRL in \framework{} comprises random and hindsight stages, our evaluation involves a comparative analysis with a baseline, \textsc{HER}, i.e., training the RL agent with hindsight goal generation.
\Cref{fig:rl_pretrain_goal_train} presents a comparison of the goal completion ratio across different methods on a validation dataset where goals are generated using $G_{rnd}$ and $G_{op}$. As depicted in \Cref{fig:rl_pretrain_goal_train}, our method surpasses HER by $\approx$3.4\%. 
\Cref{fig:rl_pretrain_goal_test} evaluates the generalization on unseen goals, addressing the second aspect mentioned earlier. It is noteworthy that the unseen goals are re-combinations of goals obtained with HER and $G_{llm}$. As indicated in \Cref{fig:rl_pretrain_goal_test}, our method excels over the baseline in terms of completion ratio.
\Cref{fig:degradation_reward_kil} answers the third point by comparing the use of KL-divergence regularizer for policy learning, considering changes in overall performance and the ability to eliminate enemies. Three metrics are designed for evaluation: (1) Mean basic reward per step, which indicates whether the current policy degenerates in performing basic skills per step against a well-trained non-goal policy, and intentional to emphasize the agent's immediate responsiveness over final results; (2) \#Enemies killed, representing the average number of enemies killed by the agent per episode; and (3) \#Enemies knocked down, representing the average number of enemies knocked down by the agent per episode. 

\subsection{Evaluating LLM-based Goal Generation}
We conducted evaluation of $G_{llm}$ through two comparative experiments on GPT-4-generated instruction datasets, aiming to investigate the impact of different instruction datasets and fine-tuning paradigms. The evaluation metrics employed encompass precision, recall, and F1 score.
It's worth noting that a potential issue in determining the precision of generating sub-goals that are close in semantics. For instance, associating the sub-goal ``moving speed'' values ``very fast'' versus ``fast'' may be perceived as a negative instance under precision measurement. Consequently, we argue that the generation of such sub-goals should weigh more in choosing sub-goal than determining values. Thus, we further propose three choice-based metrics: precision (choice), recall (choice), and F1 (choice).
\begin{table}[h!]
\centering
\resizebox{\columnwidth}{!}{
\begin{tabular}{@{}lllllll@{}}
\toprule
Dataset  & Precision      & \begin{tabular}[c]{@{}l@{}}Precision\\ (Choice)\end{tabular} & Recall         & \begin{tabular}[c]{@{}l@{}}Recall\\ (Choice)\end{tabular} & F1             & \begin{tabular}[c]{@{}l@{}}F1\\ (Choice)\end{tabular} \\ \midrule
HI & 0.435          & 0.611                                                        & 0.361          & 0.517                                                     & 0.395          & 0.560                                                 \\
AI     & 0.474          & 0.611                                                        & 0.419          & 0.532                                                     & 0.445          & 0.569                                                 \\
SI  & 0.444          & 0.601                                                        & 0.413          & 0.539                                                     & 0.428          & 0.568                                                 \\
RI   & 0.499          & 0.633                                                        & 0.414          & 0.526                                                     & 0.453          & 0.574                                                 \\ \midrule
ALL      & \textbf{0.555} & \textbf{0.685}                                               & \textbf{0.505} & \textbf{0.621}                                            & \textbf{0.529} & \textbf{0.652}                                        \\ \bottomrule
\end{tabular}
}
\caption{Evaluation on different datasets. ``ALL'' the proportional mixture of the four base datasets.}
\label{tab:data_type}
\end{table}
\Cref{tab:data_type} provides a comparison of five types of instruction datasets used in the multi-step fine-tuning process for $G_{llm}$. The comparison reveals that utilizing a mixture significantly outperforms individual base datasets, which indicates a mixture aids $G_{llm}$ in capturing human preferences and understanding the implications of each abstracted state, thereby enhancing goal generation.
\begin{table}[h!]
\fontsize{25pt}{30pt}\selectfont
\resizebox{\columnwidth}{!}{
\begin{tabular}{@{}lllllll@{}}
\toprule
Tuning                       & Precision      & \begin{tabular}[c]{@{}l@{}}Precision\\ (Choice)\end{tabular} & Recall         & \begin{tabular}[c]{@{}l@{}}Recall\\ (Choice)\end{tabular} & F1             & \begin{tabular}[c]{@{}l@{}}F1\\ (Choice)\end{tabular}       \\ \midrule
SFT                          & 0.547 & 0.663                                               & 0.490 & 0.602                                                     & 0.517 & 0.632      \\
CoTF                    & 0.533          & 0.652                                                        & 0.487          & 0.599                                                     & 0.509          & 0.624                       \\
CoTF $\rightarrow$ SFT &   0.555 & 0.685                                               & \textbf{0.505} & \textbf{0.621}                                            & \textbf{0.529} & \textbf{0.652}                                        \\
CoTF $\rightarrow$ SFT$\rightarrow$EFT            &   \textbf{0.590} & \textbf{0.694}                                               & 0.501 & 0.593                                            & 0.516 & 0.629                                        \\
\bottomrule
\end{tabular}
}
\caption{Evaluation on different tuning methods.}
\label{tab:cot}
\vspace{-.5em}
\end{table}
\Cref{tab:cot} compares four kinds of fine-tuning with the proposed multi-step fine-tuning, including (1) SFT: only use the target prompt without CoT data to supervised fine-tuning, which can be regarded as a baseline for a naive SFT; (2) CoTF: only CoT-assisted fine-tuning; (3) CoTF $\rightarrow$ SFT: further SFT target prompt after CoTF; (4) CoTF $\rightarrow$ SFT$\rightarrow$EFT: further ensemble fine-tuning target prompt after CoTF.
With the comparison, we conclude that CoTF and SFT can improve each other and achieve better performance.
Furthermore, ensemble fine-tuning significantly enhances precision while marginally decreasing recall, making it more suitable for generating accurate concise goals.
\begin{figure*}[th]
\centering
\subfigure[Goal completion rate ($1 \le \vert g \vert \le 7$)]{
\includegraphics[width=0.65\columnwidth]{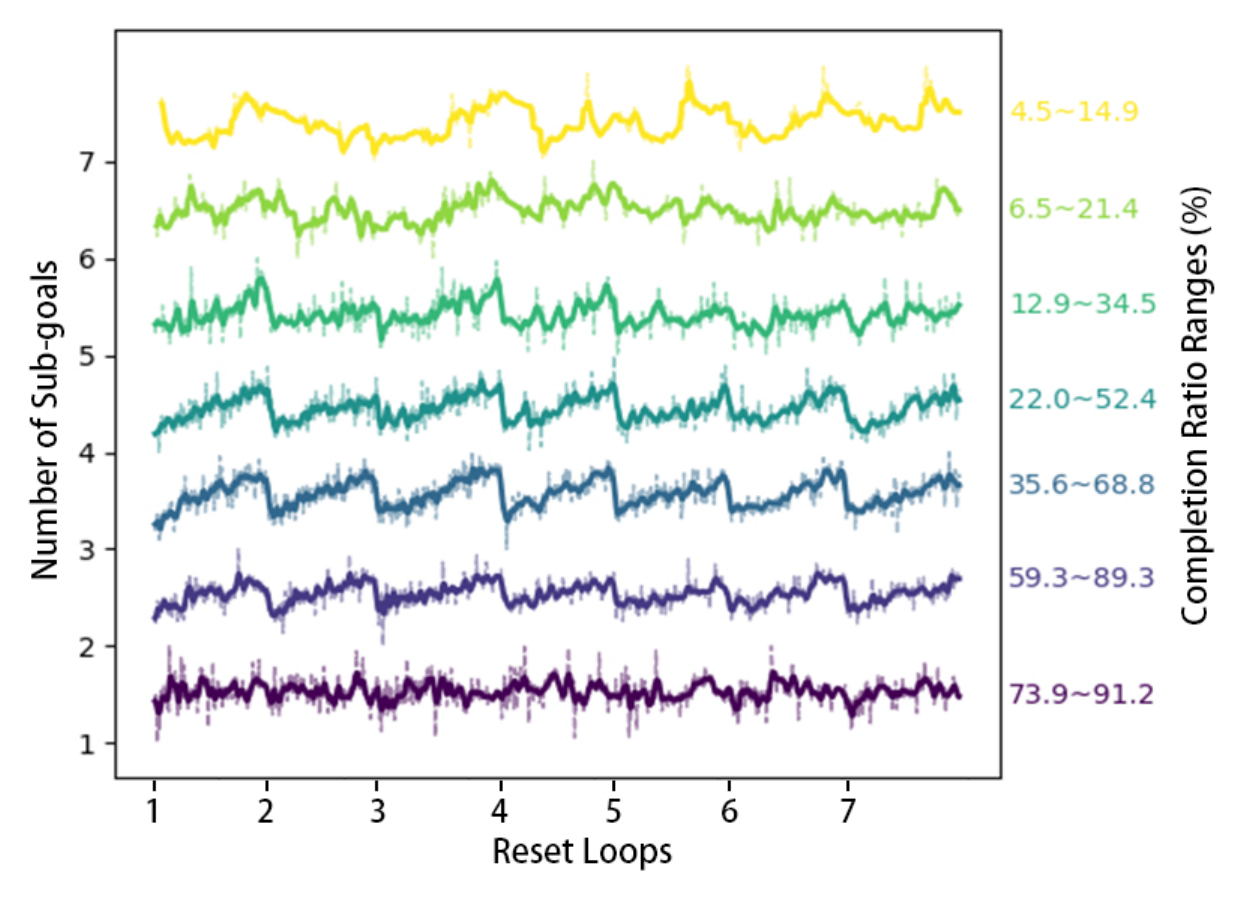}
\label{fig_co-training_reset_a}
}
\subfigure[Goal completion rate ($\vert g \vert = 3$)]{
\includegraphics[width=0.65\columnwidth]{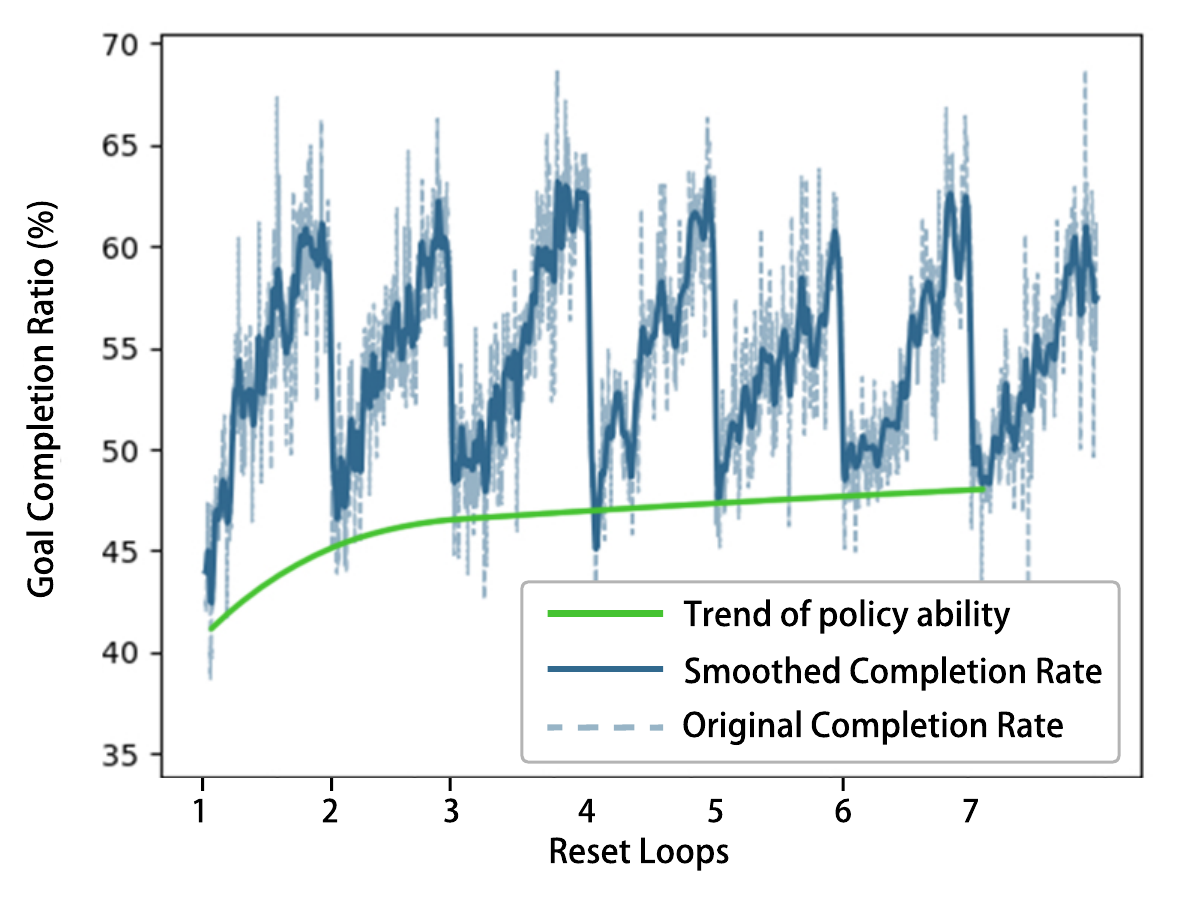}
\label{fig_co-training_reset_b}
}
\subfigure[Sub-goals distribution]{
\includegraphics[width=0.65\columnwidth]{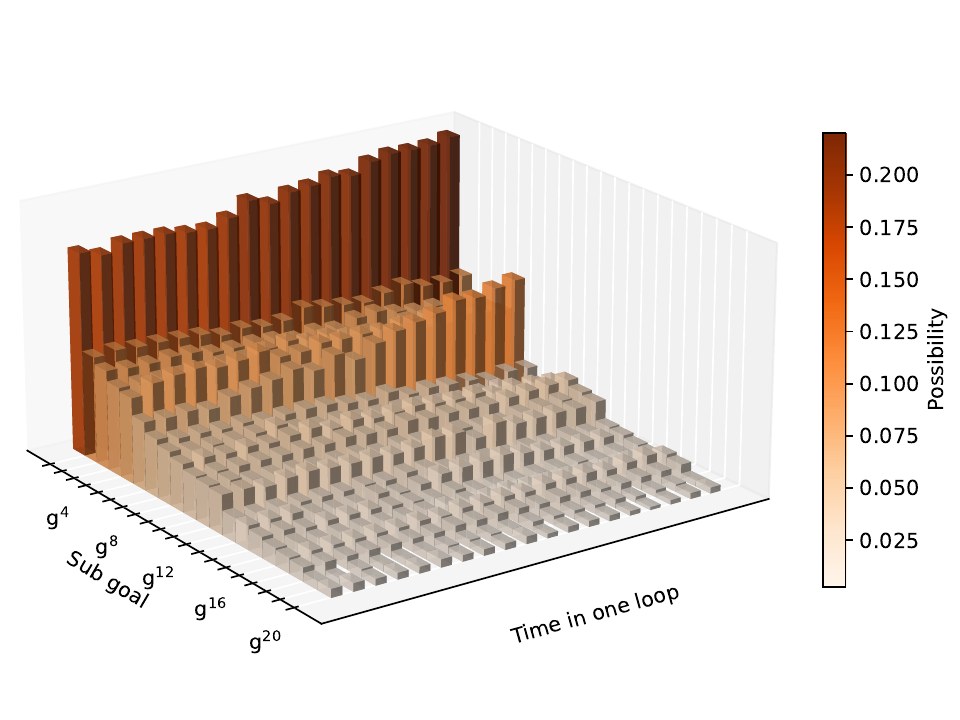}
\label{fig_co-training_reset_c}
}
\caption{(a) The completion ratio of goals with dimension size ranges from 1 to 7; (b) The goal completion ratio of goals that $|g|=3$, the trend curve reflects the improving completion ratio; (c) The sub-goals distribution changes along the training in one loop of co-training, where the description of each $g^i$ is included in \Cref{tab:frequency}.}
 \label{fig_co-training_reset}
\end{figure*}

\begin{table}[h]
\fontsize{25pt}{35pt}\selectfont
\resizebox{\columnwidth}{!}
{
\begin{tabular}{lll}
\hline
\textbf{ Instruction} & \textbf{Goal(Before co-training)} & \textbf{Goal(After co-training)}\\ \hline
\begin{tabular}[c]{@{}l@{}}Stop!\end{tabular} 
& \begin{tabular}[c]{@{}l@{}}Whether prone position: True\\ Average velocity: Static\end{tabular}                                                                      & \begin{tabular}[c]{@{}l@{}}Average velocity: Static\\Whether prone position: True\\\textcolor{cyan}{Length of distance moved: No Movement} \end{tabular}\\
\hline
\begin{tabular}[c]{@{}l@{}}Get down, \\stay hidden.\end{tabular} 
& \begin{tabular}[c]{@{}l@{}} Whether prone position: True\\ Average velocity: Static \end{tabular}
& \begin{tabular}[c]{@{}l@{}} Whether prone position: True\\ \textcolor{cyan}{Length of distance moved: No Movement}\\ Average velocity: Static \\ \textcolor{orange}{Whether seen by enemy: False} \end{tabular}\\
\hline
\begin{tabular}[c]{@{}l@{}} Enemy! \\Rush and fire.\end{tabular} 
& \begin{tabular}[c]{@{}l@{}} Whether hold a gun: True\\ Whether have bullets: True\\Horizontal direction of view: Southwest\\ Whether seen enemy: True\\Average velocity: Fast \end{tabular}
& \begin{tabular}[c]{@{}l@{}} \textcolor{cyan}{Length of distance moved: long}\\ Average velocity: Fast \\ Whether hold a gun: True\\Horizontal direction of movement: Southwest\\ Whether seen enemy: True \\\textcolor{orange}{Damage to enemy: High} \\\textcolor{cyan}{Horizontal direction of view: Southwest}\\ \end{tabular}\\
\hline
\begin{tabular}[c]{@{}l@{}} Enemies nearby, \\move to defend \\and avoid damage.\end{tabular} 
& \begin{tabular}[c]{@{}l@{}} \textcolor{pink}{Whether prone position: True}\\ \textcolor{pink}{Average velocity: Fast}\\ Whether hold a gun: True\\Health level: Full\\Whether to restore health: True\end{tabular}
& \begin{tabular}[c]{@{}l@{}} Length of distance moved: long\\ Average velocity: Fast \\ Whether prone position: False \\\textcolor{cyan}{Horizontal direction of movement: North}\\ \textcolor{cyan}{(Position of enemy: South)}\\ Whether hold a gun: True\\ \end{tabular}\\
\hline
\end{tabular}
}
\caption{Comparison of goal-generation. \textcolor{cyan}{Cyan} the helpful, \textcolor{pink}{pink} the conflicting, and \textcolor{orange}{orange} the critical sub-goals. It is evident that co-training enables goal-generation to avoid conflicts of sub-goals and improves reasonability by including helpful and critical sub-goals.}
\label{tab_example_co_training}
\end{table}
\subsection{Evaluating Co-training}
We conduct an analysis of the completion ratio corresponding to the number of valid sub-goals during the co-training process.
Though the dimension size of goal space achieves 68, the number of sub-goals for valid goals predominantly falls within the range of 1 to 7.
This is rational as completing a goal with an excessive number of sub-goals is exceedingly challenging for a policy, even impossibility for human to achieve.
Furthermore, \Cref{fig_co-training_reset_a} shows that the improvements mainly lies in $2 \le |g| \le 4$, because $|g| = 1$ is too easy while $|g| \ge 5$ is too hard to complete.
\Cref{fig_co-training_reset_b} shows a case of $|g|=3$ that co-training indeed improves the completion ratio as the green curve.
It is noteworthy that the performance suddenly downgrades at each reset. This phenomenon is attributed to the reset of $G_{llm}$ breaks the adaptation with $\pi_{g}$, avoiding being trapped in local optimal.
Meanwhile, the performance tends to converge, which indicates the successor loops produce a better adaptation between LLM and policy than before.
Additionally, we investigated the change in the generation probability of sub-goals (\Cref{tab:frequency}) during co-training.
Specifically, \Cref{fig_co-training_reset_c} illustrates changes within a training loop, while \Cref{tab_state_distributionB} indicates changes across loops. As training progresses, the probabilities associated with each $g^i$ undergo gradual modifications. For instance, sub-goals with growing probabilities are central to the agent private states due to their relatively attainable nature and influence in agent interaction. Conversely, sub-goals with falling probabilities are central to other players' states, as they are not directly changed by agent actions, and $G_{llm}$ tends to generate outputs for these sub-goals only when absolutely necessary.
To investigate the impact of co-training to $G_{llm}$, we have also identified the changes of goal-generation for an instruction, as shown in \Cref{tab_example_co_training}.
Evidently, after co-training, $G_{llm}$ demonstrates its capacity to eliminate contradictory and irrational elements within the initial objectives and exhibits the ability to introduce new sub-goals, thereby rendering the overall goal more attainable, all while retaining its exceptional semantic comprehension capabilities.

\section{Conclusion}
In this paper, we propose \framework{} experts on learning open-ended embodied agents for human-AI interaction, excelling in achieving instruction open-endedness through a two-stage learning process.
The empirical results on \Contra{} represent that \framework{} shows the potential as a practical solution for human-AI interaction in complex situations.
Despite the positive results, we admit there are still some limitations to our work that would be expected to be researched in the future—for instance, a truly open-ended goal description instead of the handcrafted goal space in the current version; supporting multi-modality input/output to free from expensive feature engineering.


\section*{Author Contribution Statement}
The authors confirm their contribution as follows:

\textbf{Shaopeng Zhai}: team leadership, open-ended learning, LLM/RLAF training, agent analysis, architecture design\\
\textbf{Jie Wang}: infrastructure/framework engineering, non-goal agent training, open-ended learning, ablation studies, feature engineering\\
\textbf{Tianyi Zhang}: non-goal agent training, open-ended learning, feature engineering\\
\textbf{Fuxian Huang}: non-goal agent training, paper writing, open-ended learning\\
\textbf{Qi Zhang}: LLM training, RLAF training, paper writing, ablation studies\\
\textbf{Ming Zhou}: co-training framework, curriculum research, paper writing\\
\textbf{Jing Hou}: LLM training, paper writing
\bibliography{ref}
\bibliographystyle{icml2024}

\appendix
\onecolumn
\section{\Contra{}: The Environment}
\label{app:env}
\Contra{} seamlessly merges the last-man-standing gameplay dynamics with the survival, exploration, and scavenging elements inherent in first-person shooting games~\citep{gautam2021battle}. The game unfolds with multiple hostile teams, necessitating players to collaborate with teammates, withstand adversaries, and strive to outlast others in the ever-changing arena. The agent's objectives encompass individual survival and the elimination of encountered enemies. 
In Contra, the agent's action interface is designed to mirror human capabilities, encompassing basic movements and actions like moving and shooting, with action intervals around 200ms, similar to the frequency of human operations. Through these actions, an agent in \Contra{} mandates a sequential acquisition of skills, starting from fundamental abilities like walking, jumping, running, and item collection. As the learning proceeds, an agent must master more intricate skills such as evading enemy projectiles and coordinating tactics with teammates. 
This characteristic defines an open-ended learning process where the agent continually explores the game environment to refine mastered skills and acquire new ones.

\paragraph{Observation Space.}
\label{sec_observation_space}
The observation space encompasses various factors, comprising unit features delineating the agent's status, as well as that of other players. Additionally, it includes environmental features characterizing interaction events and an agent-centric RGB bird's-eye-view of the local observation. For the details, we include them in \Cref{tab:features_summary}. Given the heterogeneity in the shapes and data types of these features, we adopt independent feature processing for each of them, subsequently concatenating them to serve as input for the policy and value networks.
\begin{figure}[h]
	\centering
	\includegraphics[width=.9\textwidth]{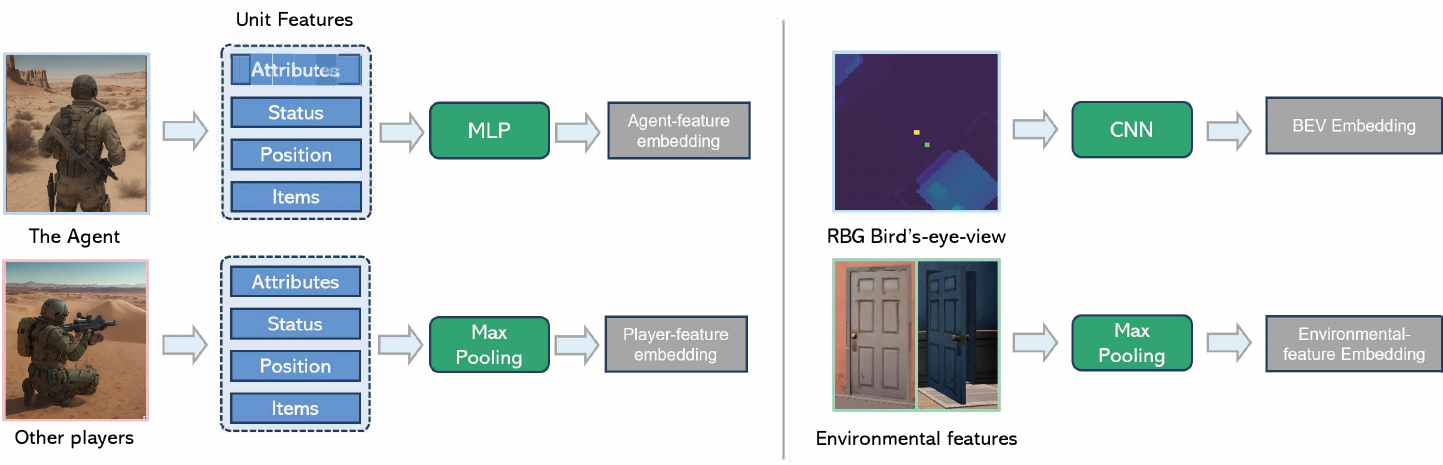}	
	\centering
	\caption{Preprocessing for an observation with four types of features.}
	\label{fig:observation_process}
\end{figure} 
In \Cref{fig:observation_process}, we present the foundational process for handling an observation. Each observation consists of four types of feature, and each type is associated with an independent network dedicated to processing and outputting the corresponding embedding.

\paragraph{Action Space.}
As introduced in \Cref{tab:action_space}, the instantiation of action space is achieved through the utilization of the micro-operation API within \Contra{}. This process gives rise to a compilation of actions characterized by diverse levels of granularity. In a detailed breakdown, the action space comprises several distinct types, namely (1) \textbf{movement direction action space}, it provides 16 discrete choices, each evenly distributed across a $360^{\circ}$ spectrum, (2) \textbf{camera yaw direction action space}, it offers 16 choices with an equitable division of $360^{\circ}$, (3) \textbf{camera pitch direction action space}, it encompasses three distinct values: $-45^{\circ}$, $0^{\circ}$, $45^{\circ}$, (4) \textbf{body action space}, it incorporates nine values: slide, stop, crouch, run, jump, ground, open or close door, rescue, and no-op and (5) \textbf{attacking action space}, it comprises: fire, reloading, treat, pick up supply, drop supply, stop and fire, stop adjust and fire, and (6) \textbf{weapon switch}, it manifests three values: use weapon 0, use weapon 1, and no-weapon. The aggregated dimensionality of the action space is quantified at 54 in total.

\paragraph{Reward Engineering.}
\label{appendix_reward}
The primary objective of our training regimen is to equip the agent with the ability to play with other players in \Contra{} while concurrently optimizing its strength in eliminating opponents. To achieve this objective, we have formulated a diverse array of rewards designed to guide policy learning. However, the complexity involved in designing and fine-tuning these rewards is evident. To simplify the engineering, our design is characterized allocating weights based on the expected value of each reward, ensuring a proportionate influence on the learning process.
In accordance with the principle, we assume a referenced maximum return of 20, with different rewards assigned proportions based on their relative importance. Specifically, for critical actions such as knocking down or killing an enemy, their values are set to approximately 4 (20\% out of 20). Conversely, for less critical actions like scouting or incurring time penalties, their values are set to less than 1 (5\% out of 20).
Detailed specifications are outlined in \Cref{tab:reward_weight}.
In accordance with the aforementioned principles, we can now construct the reward function $r(s,a)$ by linearly composing these factors, facilitating their collaborative influence on guiding policy learning. As delineated in \Cref{alg:rl}, these factors are broadly classified into three categories: basic rewards $r^{b}$, obstacle avoidance rewards $r^{oa}$, and goal achievement reward $r^{g}$. The basic rewards are primarily directed at steering the agent towards enhanced game-playing performance, encompassing collaborative engagement with teammates and eliminating adversaries, etc.
\begin{wraptable}{rh}{.7\textwidth}
    \centering
\resizebox{\linewidth}{!}{
    \begin{tabular}{llll}
        \toprule
         \textbf{Feature Class} & \textbf{Field} & \textbf{Description} & \textbf{Dimension} \\
         \midrule
         1.Unit feature & Scalar & Includes heroes, teammates, enemies, monster  &  527 \\
         \midrule
          & Attribute & Character ID,team ID,size,skills  & 28 \\
          & Status & HP,  oxygen,speed, peek type, alive state, body state,etc. & 44 \\
         Heroes& Pose & Position, rotation, camera position, camera rotation, etc. & 76 \\
          & Item & Backpack,weapon & 144 \\
         \midrule
          & Attribute & Character ID,team ID,size,skills  & 28 \\
          Teammates & Status & HP,  oxygen,speed, peek type, alive state, body state,etc. & 30 \\
         & Pose & Position, rotation, camera position, camera rotation, etc. & 43 \\
         \midrule   
          & Attribute & Character ID,team ID,size,skills  & 28 \\
          Enemies & Status & HP,  oxygen,speed, peek type, alive state, body state,etc. & 33 \\
         & Pose & Position, rotation, camera position, camera rotation, etc. & 43 \\
         \midrule     
          & Attribute & Monster type,size  & 8 \\
          Monsters& Status & HP, max HP, HP percent, target type & 6 \\
         & Pose & Position, rotation, relative position, distance  & 16 \\
         \midrule              
         2.Global feature & Scalar & Includes circle, event, door and supply  &  65 \\
         \midrule         
          & Status & State,pain,radius & 4 \\
         Circle& Position & Blue circle, white circle & 6  \\
         & Time & existence time, rest time, total time, delay time, appear time & 5 \\
         \midrule         
          & Attribute & Type,damage,elapsed time& 8 \\
         Event& Position & Occurred position & 3  \\  
         \midrule         
          & Attribute & Type,state & 5 \\
         Door& Pose & Position,relative position,rotation & 8  \\  
         \midrule           
          & Status & Type,size,quantity,etc. & 19 \\
         Supply& Position & Position,relative position,distance & 7  \\      
         \midrule
         3.Invisible enemy feature & Scalar & Invisible (nearby) enemy feature only for value estimation  &  104 \\
         \midrule           
          & Attribute & Character ID,team ID,size,skills  & 28 \\
          Invisible enemies & Status & HP,oxygen,speed, peek type, alive state, body state,etc. & 33 \\
         & Pose & Position, rotation, camera position, camera rotation, etc. & 43 \\
        \midrule
         4.Spatial feature & Scalar & BEV  &  12288 \\
         \midrule              
         BEV& Region & Altitude map and aerial view map &  3x64x64\\
         \bottomrule
    \end{tabular}
}
    \caption{The details of features in the observation space.}
    \label{tab:features_summary}
\end{wraptable}
In the case of $r^{oa}$, the objective is to promote natural navigation and forestall the agent from encountering obstacles, such as stones and trees. Regarding the implementation, penalties are imposed on the agent for deviations from the optimal path. This optimal trajectory is determined by assessing the cosine similarity between the agent's current movement direction, a 2D unit vector, provided as an environmental signal, and the expected obstacle-free trajectory derived from the action sequence in the trajectory:
\begin{equation}
    r_t^{oa}= \frac{\mathbf{d}^{env}_t \cdot \mathbf{d}^{\star}_t}{\lVert \mathbf{d}^{env}_t\rVert_2 \ast \lVert \mathbf{d}^{\star}_t\rVert_2}-1,
\end{equation}
where $\mathbf{d}^{env}_t$ the actual movement direction of the agent that can be directly accessed from the environment feedback, $\mathbf{d}^{\star}_t$ the expected movement direction, which is derived by combining the expected movement direction from the previous moment with the movement action taken at the current moment. For the convenience, we summarize the corresponding pseudo-code in \Cref{alg:ob_reward_calc}.
\begin{algorithm}
	\caption{\textsc{Calculation of $\mathbf{d}^{\star}_t$}}
	\label{alg:ob_reward_calc}
        \renewcommand{\algorithmicrequire}{\textbf{Input:}}
        \renewcommand{\algorithmicensure}{\textbf{Output:}}
	\begin{algorithmic}[1]
            \REQUIRE {$\mathbf{d}^{\star}_{t-1}$, $a_t$}  
            \ENSURE {$\mathbf{d}^{\star}_t$}    
            \IF {$a_t \in$ movement direction action space}
               \STATE $\mathbf{d}^{\star}_t \gets \begin{bmatrix}
                                                    \cos(a_t) \\
                                                    \sin(a_t) \\
                                                    \end{bmatrix}$
            \ELSIF{$a_t \in$ yaw direction action space}
                \STATE \STATE $\Delta_t = a_t - \omega$
                \STATE $\mathbf{d}^{\star}_t \gets \begin{bmatrix} \cos(\Delta_t) & -\sin(\Delta_t) \\ \sin(\Delta_t) & \cos(\Delta_t) \end{bmatrix} \mathbf{d}^{\star}_{t-1}$
            \ELSIF{$a_t \in$ \{stop; rescue; stop and fire; stop adjust and fire \}}
                \STATE $\mathbf{d}^{\star}_t \gets \begin{bmatrix}
                                                    0 \\
                                                    0 \\
                                                    \end{bmatrix}$
            \ELSE
                \STATE $\mathbf{d}^{\star}_t \gets \mathbf{d}^{\star}_{t-1} $
            \ENDIF
	\end{algorithmic}
\end{algorithm}
For the line 5 in \Cref{alg:ob_reward_calc}, we use $\Delta_t$ to denote the shifting from agent movement direction action $a_t$ to the camera vision degree $\omega$.
To address the issue of the agent getting stuck on obstacles due to short-term action sequences, we employ a smaller $\gamma$ for the corresponding value head. Specifically, we set $\gamma$ to 0.92. This adjustment helps mitigate the impact of the obstacle avoidance reward on long-term credit assignment, allowing for a more balanced consideration of immediate and future rewards in the agent's decision-making process.
\begin{table}[h]
    \centering
    \begin{tabular}{|l|l|l|}
        \hline
        \textbf{Feature} & \textbf{Weight} & \textbf{Description}   \\
        \hline
        enemy discovery & 0.02 & reward for see an enemy \\
        \hline
        detected by enemy & -0.002 & punishment for being seen by an enemy \\
        \hline
        scout & 0.0001 & reward for search for an enemy  \\
        \hline
        no-op  & -0.0002 & punishment for stopping and doing nothing  \\
        \hline
        bullet & 0.015 & reward for using and refilling bullets \\
        \hline
        health point & 0.03 & reward for health point changes\\
        \hline
        be knocked down & -2.5 & punishment for being knocked down \\
        \hline
        dead & -3.5 & punishment for being killed \\
        \hline
        damage enemy & 0.1 & reward for damaging an enemy   \\
        \hline
        knock down enemy  & 4.5 & reward for knocking down an enemy \\
        \hline
        kill enemy & 3.5 & reward for killing an enemy \\
        \hline
        approach a downed teammate & 0.001 & reward for approaching a downed teammate\\ 
        \hline
        help a downed teammate up & 0.8 & reward for helping up a downed teammate \\
        \hline
        not save a downed teammate & -0.5 & punishment for not saving a downed teammate \\
        \hline
        go to blue circle & 0.00015 & reward for going to blue circle   \\
        \hline
        be in white circle  & -0.00005 & small punishment for being outside of white circle \\
        \hline
        outside blue circle  & -0.012 & punishment for being outside of blue circle \\
        \hline
        teammate damage enemy & 0.03 & reward from teammate damaging enemies\\
        \hline
        teammate get up & 0.6 & reward from teammate getting up \\
        \hline
        I help teammate up & 4 & reward for helping teammate up \\
        \hline
        interrupt helping teammate up  & -0.05 & punishment for the interruption to help teammate up   \\
        \hline
         obstacle avoidance & 0.012 & punishment for being stuck \\
        \hline
        goal & 1 & reward of goal completion \\
        \hline
        
    \end{tabular}
    \caption{The introduction of different rewards.  }
    \label{tab:reward_weight}
\end{table}
As for the goal-achieving reward, we've introduced in the main text, please refer to \Cref{sec:stage_cgrl}.

\section{Environmental State Abstraction and Goal Space}\label{appendix:environment_state_abstraction}
For a comprehensive understanding of the game environment, a language model undergoes a fine-tuning process due to the scarcity of textual information within the simulation environment.
The need arises to articulate non-linguistic elements, and the interaction between an agent and the simulation environment is centered on continuously observing the environment's state and generating corresponding actions.
Therefore, the key aspects requiring verbalization primarily involve the state and actions of the agent.
However, given the abundance of possible observation states in the simulation environment, it is impractical to use all of these states directly as prompts for the language model, especially considering token limitations. Consequently, there is a crucial need to extract and linguistically transform the most significant meta-states to facilitate successful model interaction.
It is noteworthy that smaller language models have limitations in comprehending and manipulating numerical values effectively. To address this challenge, a deliberate effort is made to minimize the use of numerical values during the environmental linguistics process.
\begin{wraptable}{r}{.3\textwidth}
    \centering
    \begin{tabular}{|l|l|}
        \hline
        \textbf{Sub Action Space} & \textbf{Dim Size} \\
        \hline
        movement direction & 16\\
        \hline
        yaw direction & 16  \\
        \hline
        pitch direction & 3    \\
        \hline
        body action & 9 \\
        \hline
        basic action & 7 \\
        \hline
        switch weapon action & 3\\          
        \hline
    \end{tabular}
    \caption{Action space.}
    \label{tab:action_space}
\vspace{-1em}
\end{wraptable}
For example, instead of specifying an agent's speed with specific numeric metrics like ``speed: 1m/s $\rightarrow$ 3m/s'' a qualitative representation such as ``speed: slower $\rightarrow$ faster'' is adopted. This technique transforms the original continuous state into a limited, discrete meta-state, thereby enhancing the language model's understanding.
\begin{table}[h]
    \centering
    \resizebox{\columnwidth}{!}{
    \begin{tabular}{|l|l|}
        \hline
        \textbf{Sub-goal Class} & \textbf{Candidates} \\
        \hline
        Damage to enemy & [Zero,Low,Little low,Medium,Little high,High]\\
        \hline
        Whether knock down enemy & [True,False]  \\
        \hline
        Whether kill enemy & [True,False]    \\
        \hline
        Whether seen enemy & [True,False] \\
        \hline
        Whether seen by enemy  & [True,False] \\
        \hline
        Number of enemies have ever seen & [0,1,2,3,4,5,6,7,8,9,10,11,12,13,14,15]\\          
        \hline
        Length of distance moved & [No movement,Short,Medium,Long,Very long]  \\
        \hline
        Average velocity & [Static,Slow,Medium,Fast,Falling]   \\
        \hline
        Horizontal direction of movement & [West,Northwest,North,NorthEast,East,Southeast,South,Southwest] \\
        \hline
        Horizontal direction of view  & [West,Northwest,North,NorthEast,East,Southeast,South,Southwest] \\
        \hline  
        Pitch direction of view & [Low,Little low,Medium,Little high,High]\\
        \hline
        Health level & [Empty,Low,Medium,High,Full]  \\
        \hline
        Whether to restore health & [True,False]    \\
        \hline
        Whether the health is damaged & [True,False] \\
        \hline
        Whether rescued teammate  & [True,False] \\
        \hline       
        Whether be knocked down & [True,False]\\
        \hline
        Whether prone position & [True,False]  \\
        \hline
        Whether have a gun & [True,False]    \\
        \hline
        Whether have bullets & [True,False] \\
        \hline
        Whether have medical kits  & [True,False] \\
        \hline     
        Distance with nearest enemy & [Touch,Nearby,Moderate,Far,Out of reach,Extreme Far]\\
        \hline
        Whether closer with nearest enemy & [True,False]  \\
        \hline
        Whether crouch position & [True,False]    \\
        \hline
        Whether hold a gun & [True,False] \\
        \hline
        Length of distance from agent to teammate  & [Touch,Nearby,Moderate,Far,Out of reach,Extreme Far] \\
        \hline     
        Whether seen by teammate & [True,False]\\
        \hline
        Teammate's position relative to agent & [West,Northwest,North,NorthEast,East,Southeast,South,Southwest] \\
        \hline
        Whether follow with the views of teammate & [True,False]    \\
        \hline
        Whether target the same enemy as teammate & [True,False] \\
        \hline
        Whether follow with the movement direction of teammate  & [True,False] \\
        \hline 
        Horizontal direction of movement of enemy & [West,Northwest,North,NorthEast,East,Southeast,South,Southwest,None] \\
        \hline
        Velocity of enemy & [Static,Slow,Medium,Fast,Falling,None]   \\        
        \hline     
        Enemy's position relative to agent & [West,Northwest,North,NorthEast,East,Southeast,South,Southwest,None] \\
        \hline           
    \end{tabular}
    }
    \caption{Overview of sub-goal classes, we show a part of them here.}
    \label{tab:goal_space}
\end{table}
Similarly, for expediting language model understanding, a discrete action space is adopted, with each action accompanied by a succinct artificial language description. This discreet articulation of actions contributes to the overall interpretability of the language model within the simulation environment. We list the details in \Cref{tab:goal_space}.

\section{Policy Network}
\label{app:rl_network}

\begin{figure}[h]
	\centering
	\includegraphics[width=\textwidth]{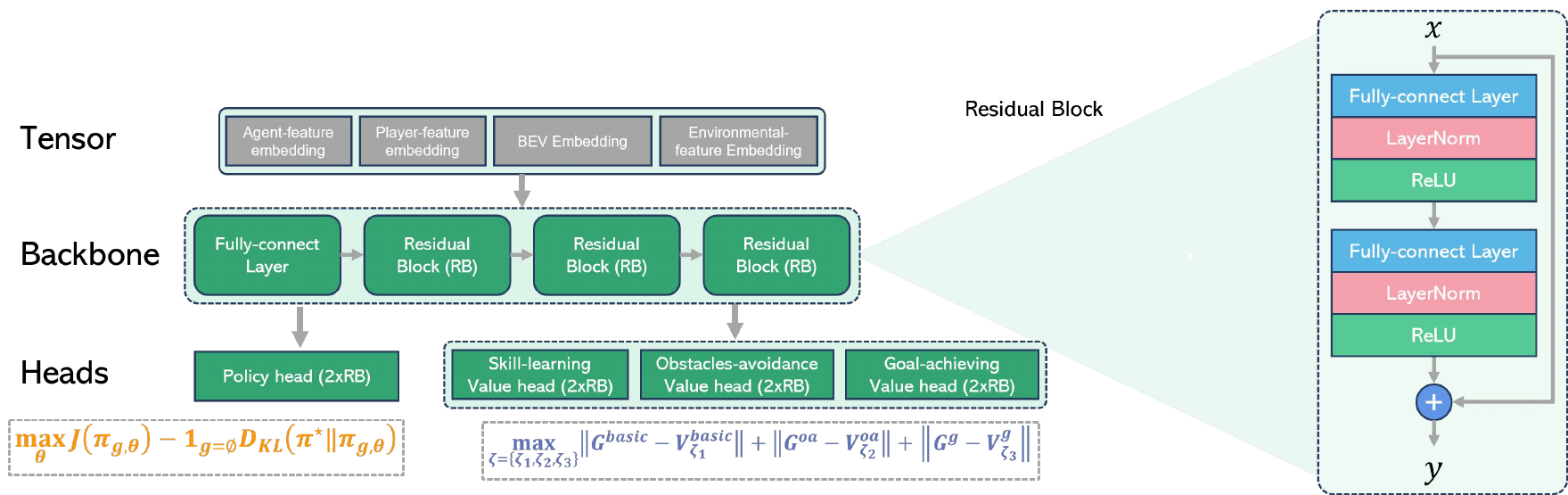}	
	\centering
	\caption{Network structure of our proposed policy. }
	\label{fig:rl_network}
\end{figure}

\begin{wraptable}{rh}{.55\textwidth}
    \centering
\resizebox{\linewidth}{!}{
    \begin{tabular}{cccl}
        \toprule
         \textbf{Date} & \textbf{Iteration} & \textbf{$\#$params} & \textbf{Change} \\
         \midrule
         4/14/2023 & 1 & 1802702 & Experiment started \\
         4/27/2023& 1808552 & 1802702 & Env-init: Random weapons \\
         5/8/2023& 2829170 & 1803087 & Action: Add a fire action  for long distance \\
         5/10/2023& 3034011 & 1803087 & Env-init:Random safe area in the whole map  \\
         5/11/2023& 3130353 & 1803855 & Observation: Add number of remaining players in the game\\
         5/12/2023& 3198564 & 2412975 & Observation: Add BEV feature\\
         5/16/2023& 3673506 & 2418111 & Observation: Add  history rotation feature \\
         5/22/2023& 4519567 & 2418368 & Observation: Add rotation change feature \\
         5/29/2023& 5442025 & 2418368 & Reward: Add rewards for teamwork  \\
         6/2/2023& 5899503 & 2418368 & Update new  game version  \\
         6/13/2023& 7306607 & 3013409 & Network: Add obstacle avoidance reward and corresponding value head\\
         6/14/2023& 7404118 & 3015457 &  Observation: Add distance feature to nearby obstacles\\
         6/16/2023& 7628098 & 3015457 & Env-init: Player numbers per team increased to 4  \\
         6/19/2023& 7974450 & 3109267 & Action: Use attention to select target to attack\\ 
         \bottomrule
    \end{tabular}
    }
    \caption{The major changes in the training procedure.
    }
    \label{tab:surgery_events}
\end{wraptable}
\Cref{fig:rl_network} is the network architecture used for non-goal RL, corresponding to the policy $\pi$. On top of observation preprocessing in \Cref{sec_observation_space}, we introduce a backbone implemented with a fully-connected layer followed by three Residual Blocks. As for the policy head and three value heads, we implemented each of them as two connected Residual Blocks.
It is noteworthy that the invisible enemy information, such as the nearest enemy's location, has also been introduced as an input to the value estimation, for the consideration of stabilizing the policy learning~\citep{vinyals2019grandmaster}.

\paragraph{Goal Processing.}
To construct the goal-conditioned policy $\pi_g$ on top of $\pi$, we need to encode a goal $g$ that is generated from $G_{rnd}$, $G_{op}$ and $G_{llm}$. Thus, we propose a network $F$ as an encoder to achieve that.
In practice, to improve the representation, we further include other inputs besides $g$ as (1) \textit{goal\_mode\_info}: a 10-dimensional vector that indicates whether current learning is goal-conditioned, the achieved times of $g$, and the rest time to complete this goal (as we preset the maximum timestep for goal completion is 30s), short in $v_{info}$; (2) \textit{sub\_goal\_masking}: a 68-dimensional 0-1 vector that indicates which sub-goals are masked for their corresponding values are 0, short in $v_{mask}$; (3) \textit{expected\_goal\_of\_current\_state}: namely, for $s_t$ at $t$, we generate a goal $g'=\textsc{Proj}(s_t)$ which is a 68-dimensional vector as the same as $g$, short in $g'$.
Then, we form the input for $F$ as $x=(g,g',v_{info},v_{mask})$.
Considering that each item in $x$ is heterogeneous on the data type, so we transform them with independent FC layers and then follows a ResNet block for each of them. We can express this process as
\begin{equation*}
    e_g = \textsc{ResNet}(\textsc{FC}(g)),\quad e_{g'} = \textsc{ResNet}(\textsc{FC}(g')),\quad e_{info} = \textsc{ResNet}(\textsc{FC}(v_{info})),\quad e_{mask} = \textsc{ResNet}(\textsc{FC}(v_{mask})),
\end{equation*}
where $e_g$, $e_{g'}$, $e_{info}$, $e_{mask}$ are the embedding representation corresponding to each input item, and all of them are the same in dimension size.
Then, we aggregate them with computing the average of them as to get a fusion embedding as
\begin{equation*}
    e_{fusion} = (e_g + e_{g'} + e_{info} + e_{mask}) / 4.
\end{equation*}
With surgery, we now concatenate the embedding of backbone of $\pi$ with $e_{fusion}$ and fusing them via a FC layer, to form a backbone for $\pi_g$.

\section{Surgery}
As the project proceeded, \Contra{} was continuously improved to satisfy richer featured environment dynamics. However, such an iterative development poses some challenges to the research of open-ended learning in an embodied situation, as the changes in API and environment attributes will make the training be non-stationary. A popular solution to resolve this issue is the surgery introduced by \citet{openai2019dota}, which significantly reduces training time by maximizing retention of previously learned abilities. Similarly, we leverage surgery in four aspects to ensure the training adapts to the new changes, including model architecture, observation space, action space, and reward functions.
\Cref{tab:surgery_events} illustrates the main changes we conducted and the corresponding parameters. For the surgery of observation space and model architecture, we have introduced a decoupled encoding in \Cref{sec_observation_space}; for the surgery of action space, we directly extend the policy head in width to satisfy the new action space; for the reward functions, the essentials are to include the newly introduced features which can contribute to the learning, as we introduced in \Cref{appendix_reward}, a linear combination has been considered to satisfy this requirement.
In our experiment, we propose three novel surgery methods, where two for model architecture and one for observation space.
The game environment has changed several times since the training started.
The changes are mainly about adding player characters, adding player skills, adding weapon, modifying the map,  etc. 
For all these changes, the proportion of new environments in which we train our policy grows slowly from  $0\%$  to $100\%$.
In this case, the variance is relatively small and the performance would quickly recover in the new environment.
\begin{figure}[h]
	\centering
	\includegraphics[width=0.99\textwidth]{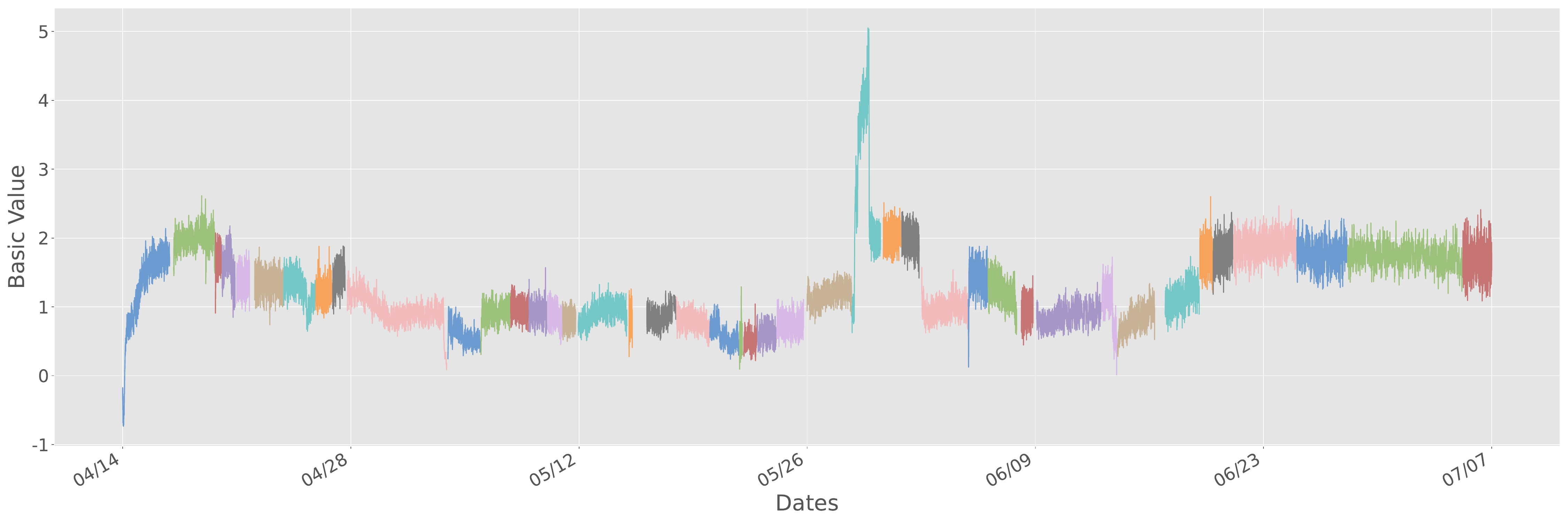}	
	\centering
	\caption{The value changes during the training process.}
	\label{fig:surgery_value}
\end{figure} 
\Cref{fig:surgery_value} evaluates the utility of surgery, illustrating the changes in basic value during the training.
It can be seen that the values change smoothly for most surgeries.
Meanwhile, the values remain stable after the surgeries. 
These results prove the effectiveness of our surgery.

\section{Datasets Construction}
\label{app:dataset_construction}
The process of fine-tuning the language model is operationalized through a question and answer paradigm. In this framework, we provide the language model with a comprehensive depiction of the present conditions pertaining to the agent, its companions, and adversaries. Additionally, we furnish the model with the linguistic instructions furnished by its teammates. Subsequently, the language model is tasked with generating the agent's subsequent target meta state in accordance with the specifics outlined in the question and answer dataset, as elucidated in \Cref{tab_prompt}.
The response is generated by GPT-4 and subsequently subjected to parsing and formatting processes facilitated by the rule-based coding mechanism. To ensure alignment between the responses generated by GPT-4 and the targeted meta-state format, a comprehensive prompt engineering endeavor was conducted to establish a set of predetermined rules for GPT-4. The rule-based prompts, which were employed to guide GPT-4's responses, are meticulously documented in \Cref{tab_rule}.

\begin{table}[h]
\centering

\begin{tabular}{|p{.9\textwidth}|}
\hline

    1. Analyze the verbal orders of teammates and players, what do you want to do? According to the command, also analysis the relevant states of teammates and enemies that need attention.\\
     The verbal command of the teammate player is `You should lie in wait', which means teammate player wants the agent to set up an ambush or take a hiding position. \\    \hline
2. Analyze which states of the agents are most relevant to the verbal commands of teammate player. The agents in the unselected states will adjust themselves to complete your plan.\\
     According to the teammate`S command:\\
     2.1. Need to hide: `Whether prone position', `Whether crouch position'\\
     2.2. Reduce movement to stay unnoticed: `Average velocity', `Length of distance moved'\\
     2.3. Ensure readiness for potential attack: `Whether hold a gun'    \\ \hline
3. Plan how these key states need to be adjusted.\\
     According to the teammate`S command:\\
     3.1. `Whether prone position': Need to hide: `False' -\textgreater{}   `True'\\
     3.2. `Whether crouch position': Alternative hiding posture if not prone: `False' -\textgreater{}   `True'\\
     3.3. `Average velocity': Reduce movement: `Fast' -\textgreater{}   `Static'\\
     3.4. `Length of distance moved': Reduce movement: `Short' -\textgreater{}   `No movement'\\
     3.5. `Whether hold a gun': Ensure readiness for potential attack: `True' -\textgreater{}   `True'\\      \hline
4. Modify the adjustment that may be wrong, and refer to the Rules to analyze which state adjustments may conflict, repeat or be unnecessary, and output the modified adjustment plan:\\
     According to the states adjustments in 3:\\
     4.1. `Whether prone position': `False' -\textgreater{} `True'\\
     4.2. `Average velocity': `Fast' -\textgreater{} `Static'\\
     4.3. `Length of distance moved': `Short' -\textgreater{} `No movement'\\
     4.4. `Whether hold a gun': `True' -\textgreater{} `True'\\
     (Removed `Whether crouch position' as it conflicts with `Whether prone position' and prone position gives better cover for an ambush.)\\    \hline
5. According to the analyze and the planing of the verbal command, further analyze the behavior tendency required in the adjustment process.\\
     Mobile: Need to stop all movement: 0\%\\
     Offense: Must be ready to attack if necessary: 20\%\\
     Waiting: The main goal is to set up an ambush and wait: 70\%\\
     Supplies: No need for supplies during this phase: 0\%\\
     Scouting: Need to observe the surroundings, but not actively scout: 10\%\\   \hline
6. Analyze how long the current command needs to be kept.\\
     According to the command of the teammate, taking a hiding position or setting up an ambush is usually done with the intention of waiting for an opportune moment to strike or observe the enemy`S movement: The current command needs to be kept by `long term'.  \\ \hline

\end{tabular}

\caption{Chain of thought response from GPT4.}
\label{tab_gpt4_response}
\end{table}

\begin{table}[h]
\centering
\begin{tabular}{|p{.9\textwidth}|}
\hline
1.Only select the most relevant and necessary states for planning, and the unplanned states will be adjusted by the agent itself                                                                                                                           \\ \hline
2.{[}Choose 1, Choose 2, ...{]} indicates the values that can be selected for the state. When you plan, you can only choose the value of the state from it, and do not invent new value not listed in {[}Choice1, Choice2, ...{]}.                         \\ \hline
3.The selected state can change the current value or maintain the current value. The agent will try to achieve and maintain the value of the state you choose after you give the plan.                                                                     \\ \hline
4.Agents don't voluntarily discard items (for example guns, bullets, medical kits) unless items are reduced or set as False in your plan, so there is no need to keep them, only to choose when making changes.                                            \\ \hline
5.Do not plan and adjust the states of teammates and enemies, they can move freely and cannot be controlled.                                                                                                                                               \\ \hline
6.Avoid conflicts of states planing. For example, agent unable to move quickly when lying down, and unable to see enemies when length of distance from agent to enemy is far away.                                                                         \\ \hline
7.Avoid the repetition of states planing. For example, if the Average velocity has been adjusted to be Fast, there is no need to adjust the Whether prone position to False, because the agent can automatically adjust state to fit overlapping meanings. \\ \hline
8.When it is necessary to refer to enemy or teammate information for planing, describe the specific state value during analysis.                                                                                                                           \\ \hline
\end{tabular}
\caption{Rule prompt for GPT4.}
\label{tab_rule}
\end{table}

In an effort to mitigate the risk of overfitting our model to concise formatted outputs, thereby preserving its capacity for environmental reasoning, we augmented our dataset with a substantial volume of Chain of Thought data. This augmentation approach entails a systematic procedure whereby the large language model is guided through a step-by-step ideation process, ultimately culminating in the attainment of the intended target state.
Concretely, our methodology commences with an initial inquiry into the semantic interpretation of the given instruction, followed by the identification of pertinent states, contemplation of state adjustments, analysis of action propensities, and an estimation of the requisite temporal considerations. Comprehensive documentation of the detailed prompts and ensuing responses derived from the Chain of Thought procedure can be found in Tables \ref{tab_gpt4_prompt} and \ref{tab_gpt4_response}.
It is noteworthy that traditional Chain of Thought processes in existing large language models often generate sequential thoughts, a method characterized by a relatively protracted temporal trajectory. This sequential reasoning approach may not be well-suited to the high real-time demands typically encountered in first-person shooter (FPS) games. Furthermore, the singular-step reasoning capabilities inherent in smaller language models are intrinsically modest and prone to errors. Consequently, the amplification of error probabilities within the Chain of Thought reasoning process may not yield superior outcomes.
In light of these considerations, we have undertaken a strategy that amalgamates Chain of Thought data with the final target state data, thereby enhancing the fine-tuning of our language model. In the course of test reasoning exercises, the language model promptly generates the ultimate target state, with the Chain of Thought information being implicitly encoded within the neural network parameters.

\paragraph{Instruction Datasets.}
\label{appendix:bi_directional_dataset_construction}
To cover a comprehensive range of instruction types and state distributions, we generated four types of instruction sets, which, when combined with states sampled from the environment, result in four different datasets. These are the HI (Human Instruction) dataset, constructed based on human-annotated commands; the SI (State Instruction) dataset, built by reverse-generating commands based on state transitions specified by the intelligent agent; the AI (Agent Instruction) dataset, constructed by main kinds of instruction which can be complete by pre-trained Agent; and the RI (Random Instruction) dataset, generated through random sampling of agent state transitions and random commands.

\begin{itemize}
    \item \emph{$\mathcal{I}_H$ (Human Instructions).} We generate open-ended instructions manually, while the corresponding states are sampled from the intelligent agent's interaction logs. These are combined and annotated using GPT-4 based on the prompting method previously described. We found that due to varying frequencies of state changes during the agent's interactions, some states are difficult to capture comprehensively only using random sampling. To ensure a more comprehensive distribution of states in the data and to facilitate better understanding by the language model, we employ a multi-round rejection sampling approach to construct state set. Let $ S $ be the set of states waiting to be sampled. We perform multiple rounds of sampling on $ S $, with $ S^{get}_{i} $ representing the set of states sampled in the $ i $-th round, initially empty. Next, we sample a state $ s $ from $ S $ without replacement and check whether $ s $ has any state values not present in $ S^{get}_{i} $. If it does, we accept it and add it to $ S^{get}_{i} $, otherwise we reject it. Once all states in $ S $ have been sampled, one round is completed. $ S^{get}_{i} $ is the result of $ i $-th round's sampling, and $ S $ will be reset for the next round. This sampling method is employed to enhance the comprehensiveness of state coverage in all datasets except the Random Instruction dataset.

    \item \emph{$\mathcal{I}_S$ (State Instructions).} We aim to cover a broader range of state changes in the instructions to enhance the language model's understanding of various state transitions. To achieve this, we design corresponding goals and instructions for all states. Specifically, for each value of each state, we generate a series of instructions that require the corresponding state and value. These are then annotated using GPT-4 based on the prompting methods previously described. The annotated results are checked; if they do not have corresponding states and values, manual annotation and modification are performed to include the relevant states.

    \item \emph{$\mathcal{I}_A$ (Agent Instructions).} We aim to initially align the planning capabilities of the language model with the pre-trained abilities of an intelligent agent based on reinforcement learning policies. To do so, we generate potential corresponding instructions based on actual state changes in agent interactions. Specifically, we first sample a series of agent state pairs at 5-second intervals. For a subset of these, we manually annotate possible corresponding instructions. We then use these manual annotations as a knowledge base and employ the "langchain" method to use these examples to guide the annotation of the remaining data using ChatGPT-3.5. Finally, we represent all the instructions as vectors using OpenAI's embedding API and perform clustering. We select the 14 most representative types of instructions and pair them cyclically with two rounds of sampled states, ultimately constructing a dataset that better reflects the fundamental execution capabilities of the intelligent agent.

    \item \emph{$\mathcal{I}_R$ (Random Instructions).} This set is primarily designed to enrich the data distribution. It is constructed by randomly generating instructions and fully randomly sampling states, and then annotated using GPT-4 based on the prompting methods previously described. 
\end{itemize}

The quantaty of the aforementioned four types of datasets is 507 for HI, 1098 for SI, 1441 for AI and 1382 for RI. Moreover, the test dataset construct instructions that differ from those used in the training data, then utilize GPT-4 to generate draft labels of goals and modified with manually filtered and annotated. This test dataset used for evaluating the model's ability to plan reasonably in response to instructions.
And the size of dataset for each tuning step is 26,568 for CoT-assited fine-tuning, 4,428 for supervised fine-tuning, and 4,994 for ensembling fine-tuning.

\section{Distributed Training Framework}
\label{app:training_system}

\begin{figure}[h]
	\centering
	\includegraphics[width=.9\textwidth]{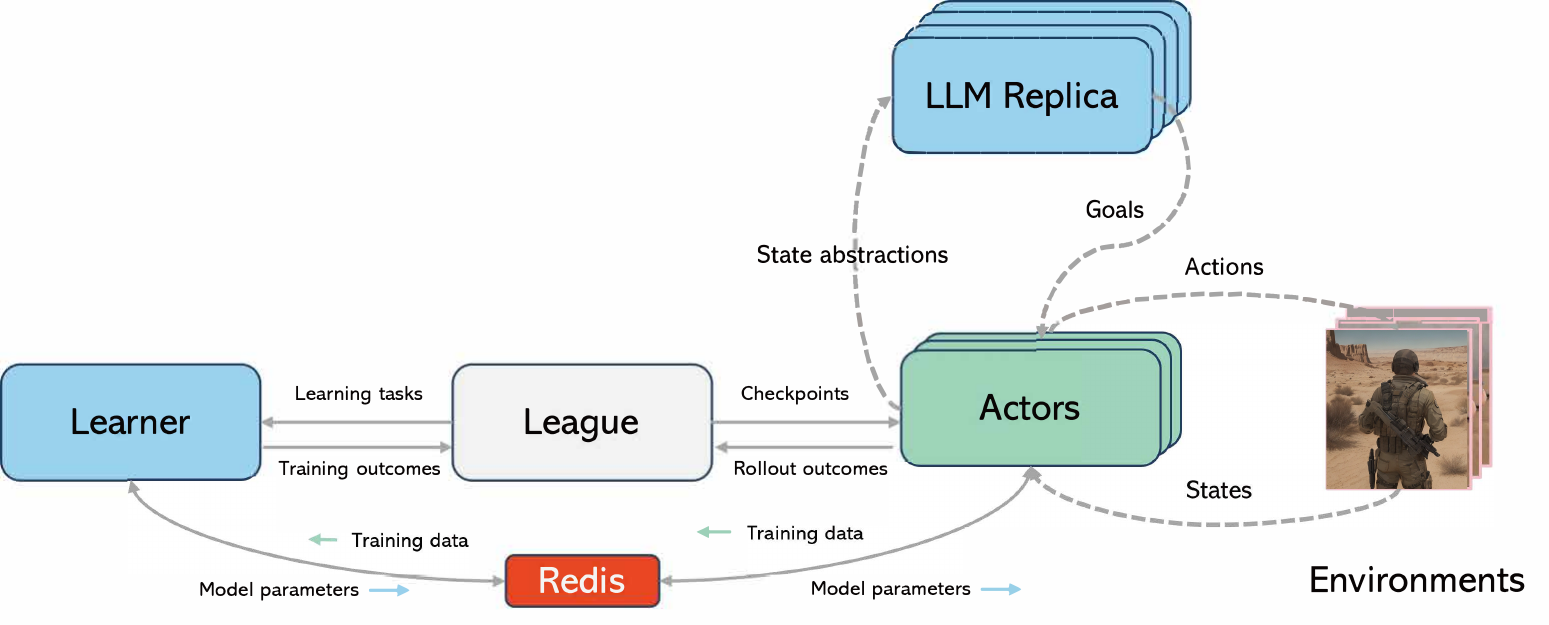}	
	\centering
	\caption{This training system has four key parts: Actor, Learner, League and LLM replicas. Actors are responsible for data collection, the Learner trains the policy model using this data, the League coordinates the overall training process and displays results, and the LLM Replicas handle goal generation and distribute them to downstream workers.}
	\label{fig:rl_framework}
\end{figure}

To improve the training efficiency, we adopt a distributed training system, shown in \Cref{fig:rl_framework}. 
In this system, the \emph{Actors} run over CPU nodes to collect training data, then send the collected data to the \emph{Learner} which is deployed on a GPU node.
We further take a LLM server to enable multiple replicas of LLM for goal generation, which improve the throughput of rollout when the RL training is switch to goal-conditioned cases.

\begin{figure}[h]
	\centering
	\includegraphics[width=0.99\textwidth]{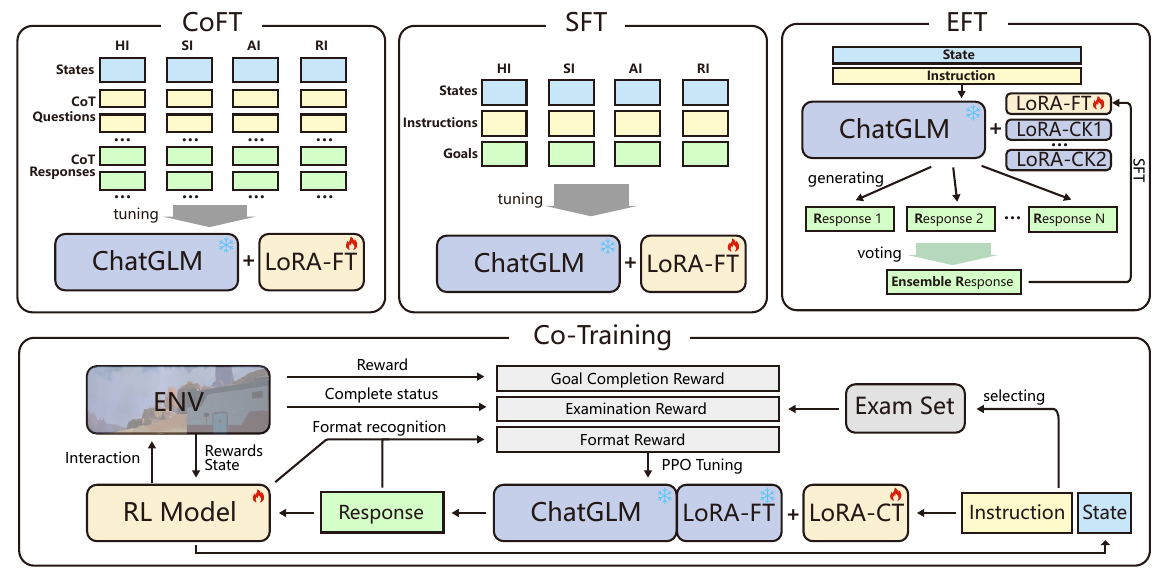}	
	\centering
	\caption{Overview of the training framework with LLM. This training framework has three kinds of LLM tuning approaches: CoFT (Chain of Thoughts assisted Fine-Tuning), SFT (Supervised Fine-Tuning), EFT (Ensemble Fine-Tuning); and one LLM-RL co-training approach.}
	\label{fig:LLM_RL_framework}
\end{figure}

\section{Parameter Settings}
Some of the hyper-parameters used in our experiment are illustrated in \cref{tab:hyper_param} and other dynamic hyper-parameters are introduced their corresponding parts.
\begin{figure}[h]
	\centering
	\includegraphics[width=\textwidth]{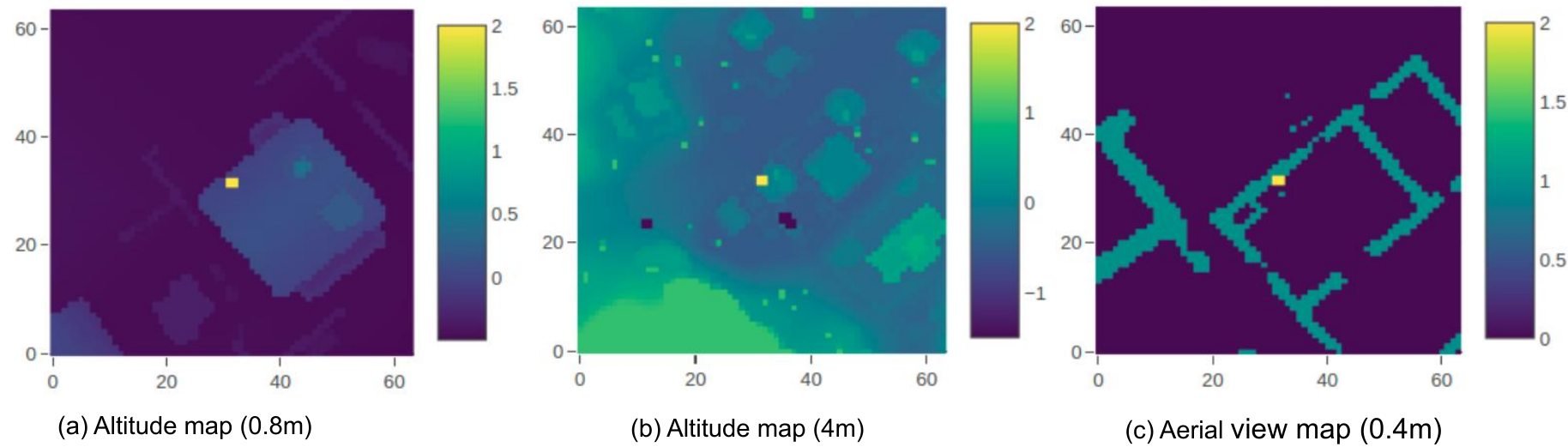}	
	\centering
	\caption{Illustration of BEV features in observation space. (a) and (b) are the altitude maps where bright areas are higher than dark areas.  (c) is the aerial view map where the disconnected areas are windows or doors. One pixel in (a), (b) and (c) denotes 0.8 meter, 4 meters and 0.4 meter respectively. The small yellow  blocks represent player positions and small blue blocks represent enemy positions. }
	\label{fig:appendix_BEV}
\end{figure} 

\section{Algorithms}
\label{appendix_algorithm}

\Cref{alg:rl} lists the pseudo-code of goal-conditioned RL procedures in Stage I,
\begin{algorithm}
	\caption{\textsc{Goal-conditioned Reinforcement Learning}}
	\label{alg:rl}
	\begin{algorithmic}[1]
        \STATE  \textbf{Input:} $\theta$ parameterizes policy $\pi$ and $\zeta = \{\zeta_1,\zeta_2,\zeta_3\}$ parameterizes value heads, goal generators $G_{rnd}$ and $G_{op}$
 
        \FOR{k=1, 2, ... }{

                \STATE Reset environment with returned initial state $s_0 \sim P(s)$
                \STATE Sample a goal: $g \sim (G_{rnd} \cup G_{op})(s_0, \Delta t, \Delta V)$
                \STATE Run policy $\pi_{\theta_{k}}$ in environment until be terminated
                \STATE Actors collect trajectories $\mathcal{D}_{\tau}$ and send them to the Learner
                \STATE Update the $\theta_k$ to $\theta_{k+1}$ with \Cref{eq:kl_policy}  
                \STATE Update $\zeta$ by $\max_{\zeta}\mathbb{E}_{s \sim \mathcal{D}_{\tau}}\left[\Vert R^{b}_t- V_{\zeta_1}^{b}(s_t)\Vert_2 + \Vert R^{oa}_t- V_{\zeta_2}^{oa}(s_t) \Vert_2 + \mathbb{1}_{g_t \neq \emptyset} \Vert R^{g}_t- V_{\zeta_3}^{g}(s_t)\Vert_2\right]$
        }
        \ENDFOR
	\end{algorithmic}
\end{algorithm}
where $R^{b}_t$, $R^{oa}_t$ and $R^{g}_t$ represent the discounted basic return, obstacle avoidance return and goal-reaching return from time step $t$ till the termination, respectively. The computation of each can be expressed as follows:
\begin{equation}
    R^{b}_t = \sum^T_{i=t}\gamma r^{b}(s_i,a_i),\quad R^{oa}_t = \sum^T_{i=t} \gamma r^{oa}(s_i,a_i), \quad R^{g}_t = \sum^T_{i=t}\gamma r^{g}(s_i,a_i,g).
\end{equation}
As for $J(\pi_{g,\theta})$ in \Cref{eq:kl_policy}, we follow the computation of policy loss in PPO to express it as
\begin{equation}
    J(\pi_{g,\theta}) = \min \left( \frac{\pi_{g,\theta}(s,a)}{\pi_{g,\theta_k}(s,a)}A^{\pi_{g,\theta_k}}(s,a), g(\epsilon, A^{\pi_{g,\theta_k}}(s,a)) \right),\text{ where } (s,a) \sim \mathcal{D}_{\tau},g(\epsilon,A) = \begin{cases}
        (1 + \epsilon) A, & A \ge 0\\
        (1 - \epsilon) A, & A > 0
    \end{cases}.
\end{equation}
$A^{\pi}(s,a)$ indicates the advantage is computed under the condition of policy $\pi$.

\section{Open-ended Goal Generation}
\begin{figure*}[th]
\centering
\subfigure[Corresponding to $(\Delta T, \Delta V)$]{
\includegraphics[width=.3\textwidth]{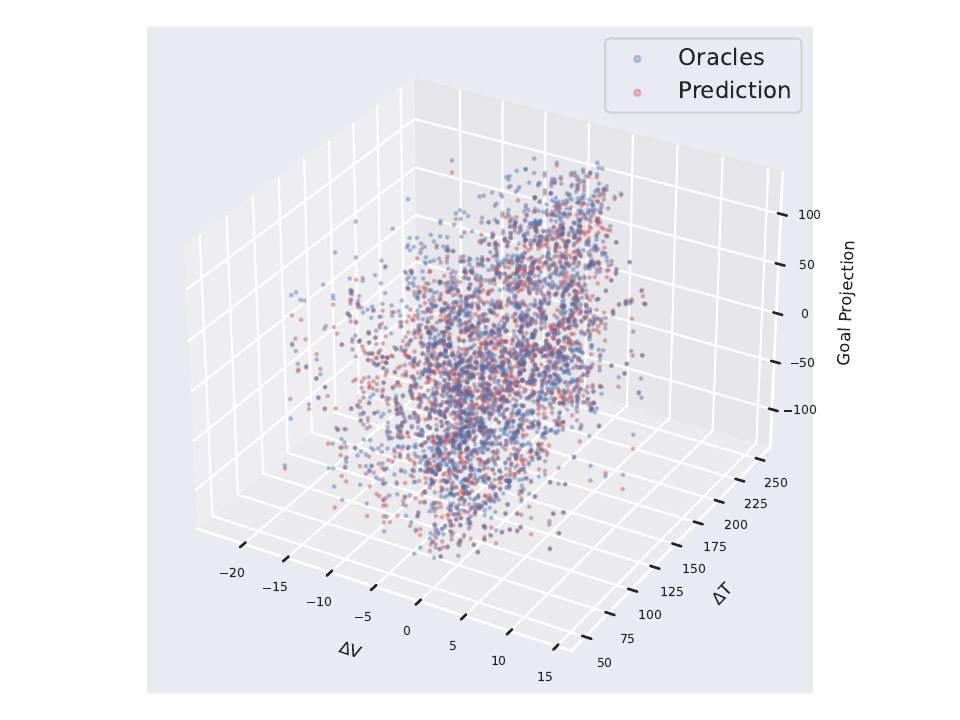}
\label{fig:goal_prediction_3d}
}
\subfigure[Corresponding to $\Delta T$]{
\includegraphics[width=.3\textwidth]{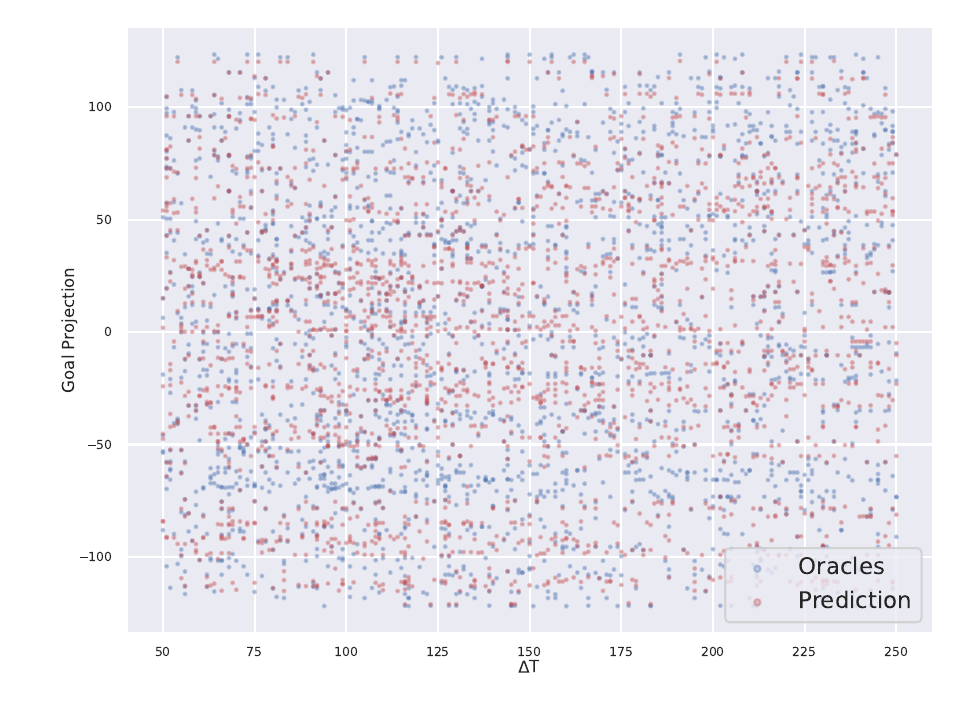}
\label{fig:goal_prediction_t}
}
\subfigure[Corresponding to $\Delta V$]{
\includegraphics[width=.3\textwidth]{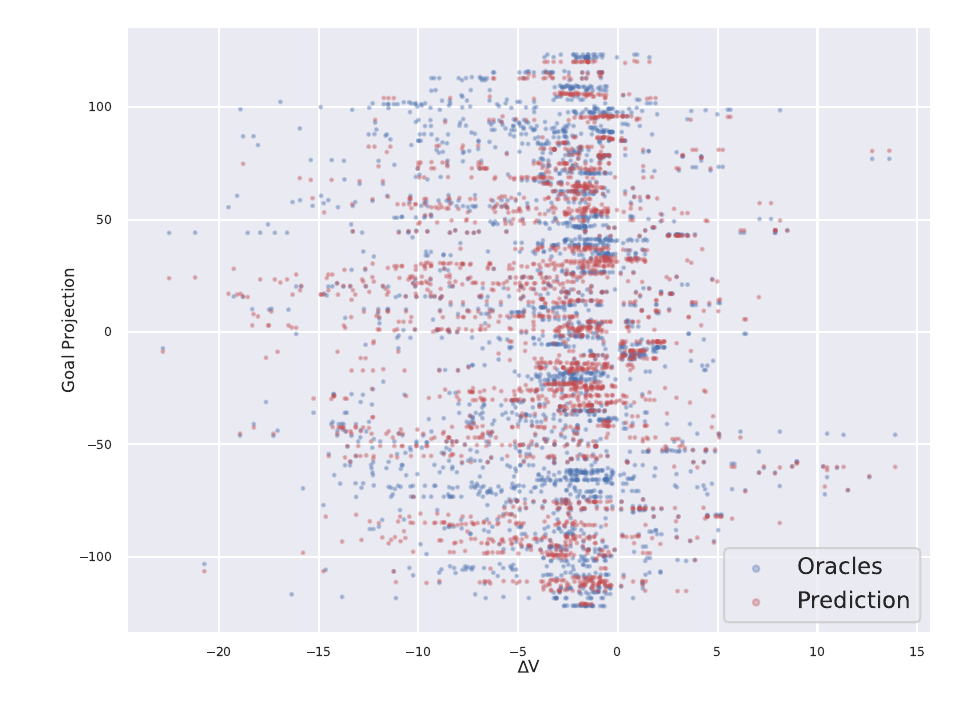}
\label{fig:goal_prediction_value}
}
\caption{Distribution comparison between real goals (Oracles) and goals generated by $G_{op}$ (Prediction). The illustration shows that $G_{op}$ generates goals that follows the real distribution, indicating good generalization on open-ended goal generation.}
\label{fig:goal_distribution_comparison}
\end{figure*}

\begin{figure}[h]
\centering
\subfigure[]{
\includegraphics[width=.45\columnwidth]{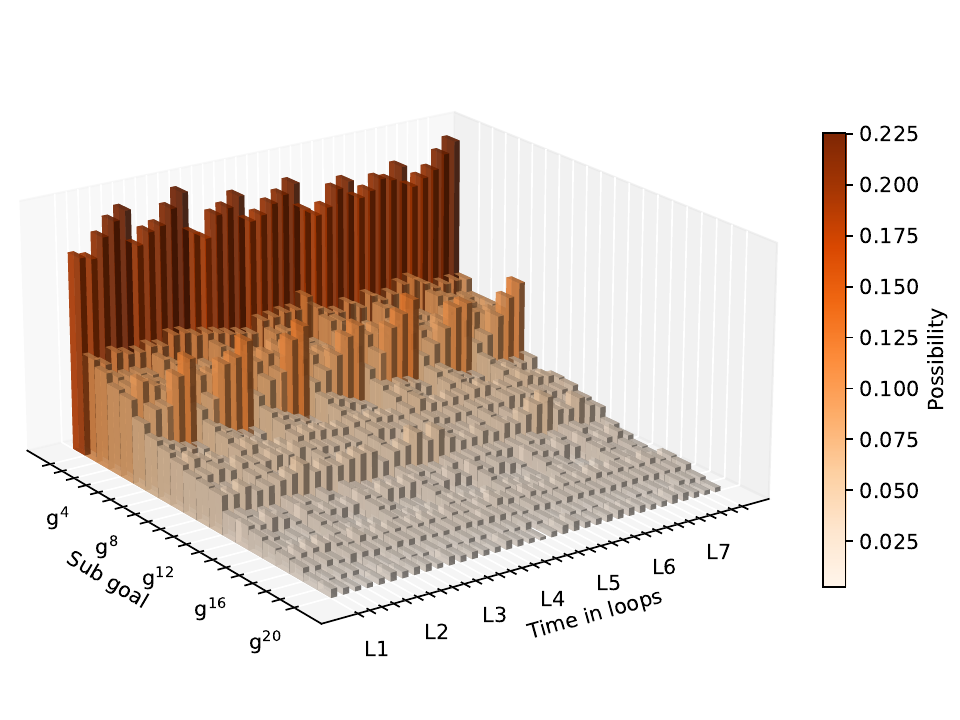}
\label{tab_state_distributionB}
}
\subfigure[]{
\includegraphics[width=.45\columnwidth]{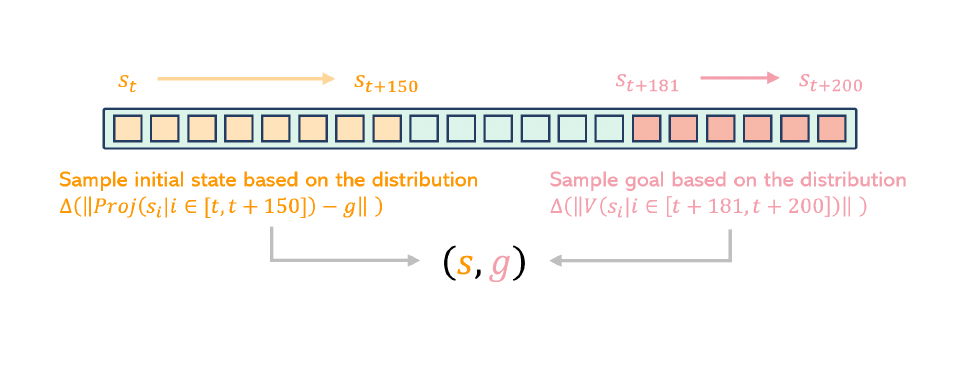}
\label{fig:hindsight_goal_generation}
}
\caption{(a) Sub-goal distribution during co-training. The 20 most frequently occurring goal meta states are filtered out and displayed. The vertical axis represents the probability of the state being output by the language model; (b) For a collected trajectory segment with length $k=200$, we firstly estimate the basic value for the last $k-j+1$ states (here $j=20$) and select one state as the goal with the probability proportional to their values.}
\end{figure}

Inspired by Hindsight Experience Replay (HER) \cite{andrychowicz2017hindsight}, we adopt a similar method to utilize the collected trajectories for learning a goal generator $G_{op}$ which accepts a state as input.
We conclude its training in two steps:
(1) constructing $(s,g)$ pairs with collected trajectories as illustrated in \Cref{fig:hindsight_goal_generation};
(2) supervised training $G_{op}$ with the above pairs and an MSE loss between the labeled goals and predicted goals.
For step (1), we split trajectories into many segments with length of 200 timesteps. Then, we randomly sample a state $s$ from the first 150 steps and sample a state $s'$ from the last 20 steps to derive a goal $g = \textsc{Proj}(s')$, with a distribution proportional to their basic value $V^{basic}(s')$.
For step (2), we train $G_{op}$ with $s$, $\Delta t$, $V^{basic}(s)$ and $V^{basic}(s')$ as input to generate goals, where $\Delta t$ the time slot of goal completion, $V^{basic}(s')$ and $V^{basic}(s)$ the basic state value.
\begin{wraptable}{rh}{.4\textwidth}
    \centering
\resizebox{\linewidth}{!}{
    \begin{tabular}{|l|l|}
        \hline
        PPO clip eps & 0.2 \\
        \hline
        Optimizer & Adam \\
        \hline
        Learning rate & 0.0001 \\
        \hline
        Batch size & 20480\\
        \hline
        Number of CPUs & 5120 (AMD EPYC 7H12 64-Core)\\
        \hline
        Number of GPUs  & 2 (A100)\\
        \hline
        $\gamma \quad (basic)$ & 0.995\\
        \hline
        $\gamma \quad (oa)$ & 0.92\\
        \hline
        $\gamma \quad (goal)$ & 0.993\\
        \hline        
        $\lambda$ & 0.95\\
        \hline
        Entropy coefficient & 0.025\\
        \hline
        Unroll length & 20\\
        \hline
        Sample max use times & 3\\
        \hline
        Gradient clip threshold & 10\\
        \hline
    \end{tabular}
    }
    \caption{Parameter settings for RL.}
    \label{tab:hyper_param}
    \vspace{-3em}
\end{wraptable}
\paragraph{Validation.}
To validate the efficiency of $G_{op}$, we conduct a comparison between generated goals and oracles w.r.t distribution, as illustrated in \Cref{fig:goal_distribution_comparison}. Empirically, we first construct a test dataset of goals (we label them as Oracles) with trajectories that are sampled with multiple checkpoints of non-goal policy, to ensure the diversity. Then, the tuples of $(s,\Delta t, \Delta V)$ that corresponding to the oracles are used to generate goal predictions (we label them as Prediction) with $G_{op}$. To visualize the goal distribution, we leverage TSNE~\citep{van2008visualizing} to shrink the goal dimension from 68 to 1. The results in \Cref{fig:goal_distribution_comparison} show that the distribution of Prediction well matches the distribution of Oracles.

\section{Ablation Study on LLMs}
\paragraph{The Impact of Lora Rank.}
We evaluate the impact of the rank parameter on performance during LoRA fine-tuning of large language model neural networks. Generally speaking, the larger the rank parameter, the more comprehensive and thorough the fine-tuning of the neural network, but the corresponding training time and model footprint will be larger.
\begin{wraptable}{rh}{.5\textwidth}
\centering
\resizebox{.5\textwidth}{!}{
\begin{tabular}{@{}cllllllll@{}}
\toprule
Rank & Precision      & \begin{tabular}[c]{@{}l@{}}Precision\\ (Choice)\end{tabular} & Recall         & \begin{tabular}[c]{@{}l@{}}Recall\\ (Choice)\end{tabular} & F1             & \begin{tabular}[c]{@{}l@{}}F1\\ (Choice)\end{tabular} & Accurate       & \begin{tabular}[c]{@{}l@{}}Accurate\\ (Choice)\end{tabular}       \\ \midrule
8         & 0.544 & 0.672                                               & 0.482          & 0.608                                                     & 0.502          & 0.629                                                 & 0.060          & 0.124                                              \\
16        & 0.550          & 0.673                                                        & 0.487          & 0.601                                                     & 0.507          & 0.626                                                 & 0.070 & 0.124                                                     \\
32        & \textbf{0.555} & \textbf{0.685}                                               & 0.505 & 0.621                                            & \textbf{0.529} & \textbf{0.652}                                        & 0.065          & \textbf{0.159}                                                    \\
64        & 0.547          & 0.675                                                        & 0.501          & 0.616                                                     & 0.519          & 0.635                                                 & 0.070 & 0.124                                                     \\
128       & 0.552          & 0.684                                                        & \textbf{0.507} & \textbf{0.626}                                            & 0.524 & 0.645                                        & \textbf{0.075}          & 0.134                                                      \\ \bottomrule
\end{tabular}
}
\caption{Evaluation on lora rank.}
\label{tab_rank}
\vspace{-2em}
\end{wraptable}
The experimental results are shown in Table \ref{tab_rank}. The size of lora rank has little impact on model performance indicators, but a large rank will cause the model training time and the size of the saved parameter file to increase dramatically.

\paragraph{The Impact of Lora Target.}
We next verified which neural networks in fine-tuning the ChatGLM-6B large language model can achieve the best performance.
The experimental results are shown in \Cref{tab_target}.
It is worth noting that only fine-tuning the MLP network without fine-tuning the attention network can achieve the best training results.
Although generally speaking, the mainstream fine-tuning task of large language models is to fine-tune the attention layer network, but that task usually focuses more on answer semantics. In our task, we pay more attention to the format to meet the metastate parsing requirements, so fine-tuning the MLP network can achieve better results.

\begin{table}[h]
\centering
\resizebox{.7\textwidth}{!}{  
\begin{tabular}{@{}lllllllll@{}}
\toprule
\textbf{Dataset}   & \textbf{Precision}      & \begin{tabular}[c]{@{}l@{}}\textbf{Precision}\\ \textbf{(Choice)}\end{tabular} & \textbf{Recall}         & \begin{tabular}[c]{@{}l@{}}\textbf{Recall}\\ \textbf{(Choice)}\end{tabular} & \textbf{F1}             & \begin{tabular}[c]{@{}l@{}}\textbf{F1}\\ \textbf{(Choice)}\end{tabular} & \textbf{Accurate}       & \begin{tabular}[c]{@{}l@{}}\textbf{Accurate}\\ \textbf{(Choice)}\end{tabular}       \\ \midrule
\textbf{Attention} & \textbf{0.555} & \textbf{0.685}                                               & \textbf{0.505} & \textbf{0.621}                                            & \textbf{0.529} & \textbf{0.652}                                        & 0.065          & \textbf{0.159}                  \\
Mlp       & 0.549 & 0.664                                               & 0.482 & 0.587                                            & 0.514 & 0.620                                        & 0.065 & 0.134                                             \\ \midrule
All       & 0.529          & 0.642                                                        & 0.471          & 0.581                                                     & 0.485          & 0.596                                                 & \textbf{0.069}          & 0.119                                                      \\ \bottomrule
\end{tabular}
}
\caption{Evaluation on LoRA target.}
\label{tab_target}
\end{table}

\begin{table}[h]
\centering
\begin{tabular}{ll}
\toprule
\textbf{Symbol} & \textbf{Sub-goal Class}                                                \\ \midrule
$g^1$    & Average velocity                                       \\
$g^2$    & Horizontal direction of movement                       \\
$g^3$   & Whether seen enemy                                     \\
$g^4$    & Whether hold a gun                                     \\
$g^5$    & Whether prone position                                 \\
$g^6$    & Length of distance moved                               \\
$g^7$    & Length of distance from agent to teammate              \\
$g^8$    & Distance with nearest enemy                            \\
$g^9$    & Whether seen by enemy                                  \\
$g^{10}$   & Damage to enemy                                        \\
$g^{11}$   & Whether have bullets                                   \\
$g^{12}$   & Horizontal direction of view                           \\
$g^{13}$   & Whether follow with the movement direction of teammate \\
$g^{14}$   & Whether crouch position                                \\
$g^{15}$   & Whether have a gun                                     \\
$g^{16}$   & Whether have medical kits                              \\
$g^{17}$   & Whether to restore health                              \\
$g^{18}$   & Health level                                           \\
$g^{19}$   & Whether knock down enemy                               \\
$g^{20}$   & Whether target the same enemy as teammate              \\ \bottomrule
\end{tabular}
\caption{Top 20 sub-goals ranked by frequency.}
\label{tab:frequency}
\end{table}

\paragraph{The Impact of Dataset Scale.}
We conduct experiments of various models with four percentages of fine-tuning train set, i.e., 100\%, 30\%, 10\%, 3\%, on the goal generation task. 
The results are shown in \Cref{tab_samll_data}.
It can be seen that as the amount of data gradually decreases, the performance of various training indicators gradually deteriorates under various settings.
However, the smaller the amount of data, the greater the improvement brought by pre-training of our proposed CoFT method.
The results show that the CoFT method we proposed can effectively collect and expand the chain of thought data related to the final goal, thereby avoiding overfitting of the training set in the case of small data.

\begin{table}[]
\centering
\begin{tabular}{llllllll}
\hline
Dataset Size           & \begin{tabular}[c]{@{}l@{}}Training\\ Method\end{tabular} & Precision      & \begin{tabular}[c]{@{}l@{}}Precision\\ (Choice)\end{tabular} & Recall         & \begin{tabular}[c]{@{}l@{}}Recall\\ (Choice)\end{tabular} & F1             & \begin{tabular}[c]{@{}l@{}}F1\\ (Choice)\end{tabular} \\ \hline
\multirow{4}{*}{100\%} & CoFT                                                      & 51.85\%        & 64.84\%                                                      & 46.68\%        & 57.91\%                                                   & 49.13\%        & 61.18\%                                               \\
                       & CoFT $\rightarrow$ SFT                                    & 55.48\%        & 68.52\%                                                      & 50.48\%        & 62.10\%                                                   & 52.86\%        & 65.15\%                                               \\
                       & SFT                                                       & 54.70\%        & 65.20\%                                                      & 49.00\%        & 60.20\%                                                   & 51.70\%        & 63.20\%                                               \\
                       & Improve Rate                                              & {\ul 1.42\%}   & {\ul 5.09\%}                                                 & {\ul 3.02\%}   & {\ul 3.16\%}                                              & {\ul 2.25\%}   & {\ul 3.09\%}                                          \\ \hline
\multirow{4}{*}{30\%}  & CoFT                                                      & 49.43\%        & 62.66\%                                                      & 44.45\%        & 56.29\%                                                   & 46.81\%        & 59.30\%                                               \\
                       & CoFT $\rightarrow$ SFT                                    & 49.92\%        & 62.59\%                                                      & 45.51\%        & 57.65\%                                                   & 47.61\%        & 60.02\%                                               \\
                       & SFT                                                       & 46.12\%        & 60.96\%                                                      & 33.68\%        & 45.39\%                                                   & 38.93\%        & 52.03\%                                               \\
                       & Improve Rate                                              & {\ul 8.25\%}   & {\ul 2.68\%}                                                 & {\ul 35.11\%}  & {\ul 27.01\%}                                             & {\ul 22.30\%}  & {\ul 15.34\%}                                         \\ \hline
\multirow{4}{*}{10\%}  & CoFT                                                      & 45.58\%        & 60.84\%                                                      & 41.77\%        & 53.92\%                                                   & 43.59\%        & 57.17\%                                               \\
                       & CoFT $\rightarrow$ SFT                                    & 48.06\%        & 61.01\%                                                      & 43.15\%        & 54.31\%                                                   & 45.47\%        & 57.47\%                                               \\
                       & SFT                                                       & 42.08\%        & 55.31\%                                                      & 30.86\%        & 41.45\%                                                   & 35.61\%        & 47.39\%                                               \\
                       & Improve Rate                                              & {\ul 14.20\%}  & {\ul 10.32\%}                                                & {\ul 39.80\%}  & {\ul 31.03\%}                                             & {\ul 27.69\%}  & {\ul 21.28\%}                                         \\ \hline
\multirow{4}{*}{3\%}   & CoFT                                                      & 39.42\%        & 58.35\%                                                      & 34.28\%        & 50.10\%                                                   & 36.67\%        & 53.91\%                                               \\
                       & CoFT $\rightarrow$ SFT                                    & 41.61\%        & 60.40\%                                                      & 34.55\%        & 50.32\%                                                   & 37.75\%        & 54.90\%                                               \\
                       & SFT                                                       & 17.66\%        & 38.28\%                                                      & 13.33\%        & 29.47\%                                                   & 15.20\%        & 33.30\%                                               \\
                       & Improve Rate                                              & {\ul 135.52\%} & {\ul 57.78\%}                                                & {\ul 159.15\%} & {\ul 70.79\%}                                             & {\ul 148.43\%} & {\ul 64.88\%}                                         \\ \hline
\end{tabular}
\caption{Language model performance evaluation with different sizes of fine-tuning training set. The underlined ``Improve Rate'' values represent the improvement percentage of the ``CoFT $\rightarrow$ SFT'' method relative to ``SFT'' method.}
\label{tab_samll_data}
\end{table}

\section{Reward Functions for Co-Training}
\label{app:cot_reward}

\paragraph{Agent Feedback Rewards.}
The calculation of the agent feedback reward is multifaceted, aiming to reflect the degree of completion as feedback for the training of the LLM. Specifically, three aspects are considered to satisfy the requirements, and the total agent feedback reward is given by the sum of them:

\begin{itemize}
    \item \textbf{$r^f_{g}$ - Minimal Distance to a Goal When Satisfying Environment Termination}. As depicted by \Cref{eq:reward_goal-oriented}, the agent progressively reduces the distance between the initial state and the goal, scaling it by the magnitude of the initial state-goal difference:
    \begin{equation}
        R^f_{g} = \sum_{t=1}^T \lVert\frac{\lvert g - Proj(s_{t-1}) \rvert - \lvert g - Proj(s_t) \rvert}{\lvert g-Proj(s_0) \rvert + \epsilon}\rVert_1,\text{ where }\epsilon = \text{1e-6}
        \label{eq:reward_goal-oriented}
    \end{equation}

    \item \textbf{$r^f_{keep}$ - Reward Indicating How Long the Goal Can Be Kept}. As depicted by \Cref{eq:reward_keep}, upon accomplishing the goal, the agent receives a reward proportional to the cumulative number of steps taken to sustain the goal state, scaled by the count of distinct sub-goals between the initial state $s_0$ and the goal $g$, i.e. $n(g \cap \textsc{Proj}(s_0))$:
    \begin{equation}
        R^f_{keep} = n(g \cap \textsc{Proj}(s_0)) \cdot \sum_{t=0}^T \mathbb{1}_{g \cap \textsc{Proj}(s_t) \ne \emptyset}
        \label{eq:reward_keep}
    \end{equation}

    \item \textbf{$r^f_{rnd}$ -   Reward Indicating Whether the Generated Goal is Reachable for the Current State}. RND~\citep{burda2018exploration,nikulin2023anti,zhang2021noveld,du2023guiding} is an effective method to measure the visiting frequency of states or transitions in RL, where higher a RND score (reward), the more frequent a state is visited. Thus, we can leverage such a method to quantify how novel a state is: 
    \begin{equation}
        R^f_{rnd} = -\sum_{t=0}^T \|\varphi(E(s_t, g))-\varphi^{\star}(E(s_t, g))\|,
    \end{equation}
    where $\varphi^{\star}$ a target network which shares the same architecture as the RND predictor but the network is non-trainable.
\end{itemize}

Thus, we express $R^f$ as $R^f = R^f_g + R^f_{keep} + R^f_{rnd}$.

\paragraph{Examination Reward Function.}
The examination reward function is introduced as an intrinsic signal to encourage the LLM to generate goals with essential sub-goals. We use the SI dataset as the examination set $\mathcal{I}_S$. For each training iteration, a batch of instructions $\mathcal{I}_{train}$ is randomly sampled from the full instruction dataset $\mathcal{I}$, and corresponding goals $g$ are generated. After the agent finishes its rollout, the examination reward for each batch is computed based on the intersection $\mathcal{I}_{\cap} = \mathcal{I}_S \cap \mathcal{I}_{train}$. For non-empty $\mathcal{I}_{\cap}$, an examination reward for each instruction in $\mathcal{I}_{\cap}$ is computed as:

   \begin{equation}
   r^e(\iota, g, g_{sub}) = 
   \begin{cases}
       +2 & g_{sub} \in g \\
       -2 & \text{otherwise}
   \end{cases},
   \forall \iota \in \mathcal{I}_{\cap}
   \end{equation}

Then, $R^e$ is calculated as $R^e = \frac{1}{\vert \mathcal{I}_{\cap} \vert}\sum_{\iota \in \mathcal{I}_{\cap}}r^e(\iota,g,g_{sub} \vert g = G_{llm}(s,\iota))$.

\paragraph{Formatting Reward Function.}
The formatting reward $R^m$ for each generated goal is calculated by computing an edit distance, utilizing the Wagner-Fischer algorithm~\citep{wagner1974string}.

With the defined reward functions, RLAF is applied with a reward function $R=R^f+R^e+R^m$ and Proximal Policy Optimization (PPO) for each data point in a batch.

\begin{figure}[h]
	\centering
	\includegraphics[width=\textwidth]{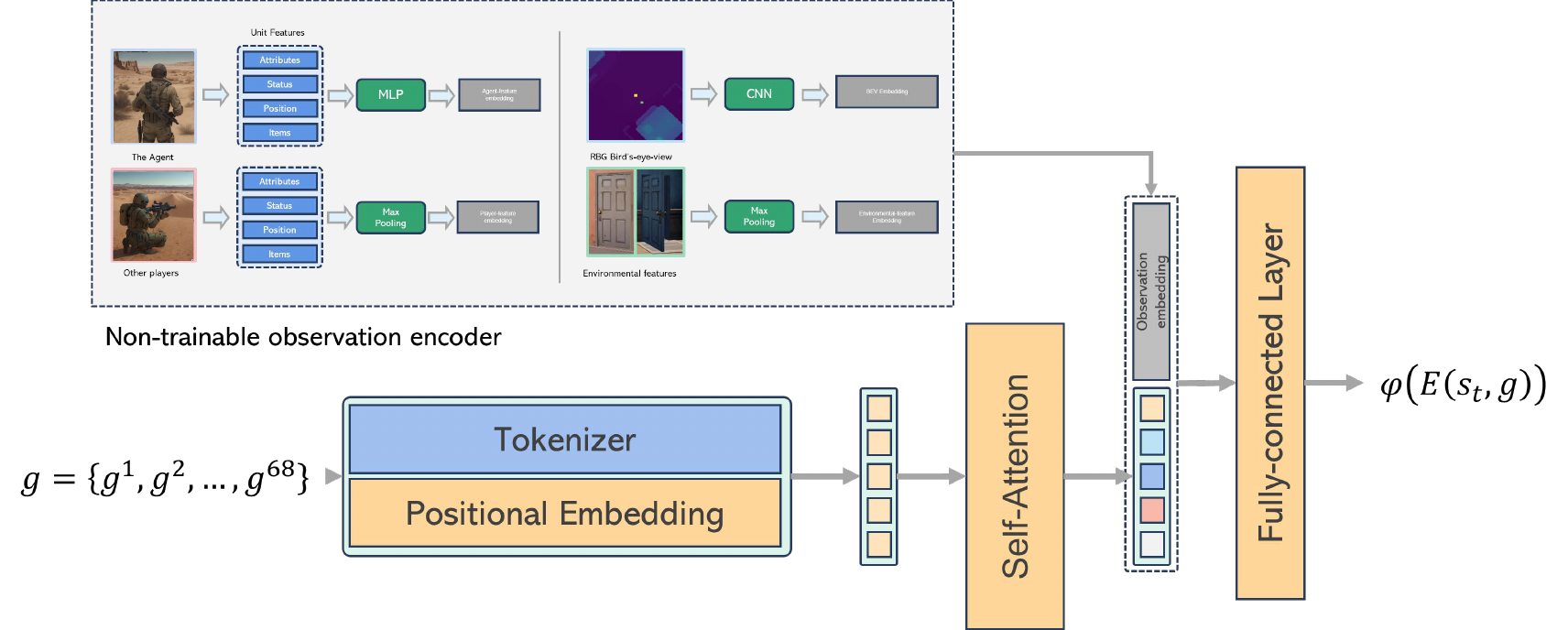}	
	\centering
	\caption{Implementation of the RND predictor network. }
	\label{fig:appendix_rnd}
\end{figure}

\begin{table}[h]
\centering
\begin{tabular}{|p{.9\textwidth}|}
\hline
In order to complete the command `You should lie in wait', let us plan the states of the agent step by step using the following template:    \\ \hline
1. Analyze the verbal orders of teammates and players, what do you want to do? According to the command, also analysis the relevant states of teammates and enemies that need attention. \\
The verbal command of the teammate player is {[}Command{]}, which means teammate player wants the agent...        \\ \hline
2. Analyze which states of the agents are most relevant to the verbal commands of teammate player. The agents in the unselected states will adjust themselves to complete your plan (analyze the reason first, then select key states one by one as few as possible and as important as possible according to the degree of importance)?  \\
According to the teammate's command:  \\
2.1. {[}Reason1{]}: {[}State1{]}  \\
2.2. {[}Reason2{]}: {[}State2{]}  \\
...                     \\ \hline
3. Plan how these key states need to be adjusted (analyze the reason first, and then make adjustments one state by one state, the state can be changed or remain the same, and must be selected from the value range of the game state {[}Choice 1, Choice 2, ...{]})?  \\
According to the teammate's command:  \\
3.1. {[}State1{]}: {[}Reason1{]}: {[}Current\_value1{]} -\textgreater {[}Target\_value2{]}  \\
3.2. {[}State2{]}: {[}Reason2{]}: {[}Current\_value1{]} -\textgreater {[}Target\_value2{]}  \\
...          \\ \hline
4. Modify the adjustment that may be wrong, and refer to the Rules to analyze which state adjustments may conflict, repeat or be unnecessary, and output the modified adjustment plan:  According to the states adjustments in 3...  \\
4.1. {[}State1{]}: {[}Current\_value1{]} -\textgreater {[}Target\_value2{]}  \\
4.2. {[}State2{]}: {[}Current\_value1{]} -\textgreater {[}Target\_value2{]}  \\
...                      \\ \hline
5. According to the analyze and the planing of the verbal command, further analyze the behavior tendency required in the adjustment process (the proportion of Mobile, Offense, Waiting, Supplies, Scouting, first analyze the reason, and then calculate the percentage)  \\
Mobile: {[}Reason1{]}: {[}Percent1{]}  \\
Offense: {[}Reason2{]}: {[}Percent2{]}  \\
Waiting: {[}Reason3{]}: {[}Percent3{]}  \\
Supplies: {[}Reason4{]}: {[}Percent4{]}  \\
Scouting: {[}Reason5{]}: {[}Percent5{]}           \\ \hline
6. Analyze how long the current command needs to be kept (for example, the command of `killing the enemy' needs to be kept for a `short term', and the command of `pay attention to reconnaissance' needs to be kept for a `long term'. First analyze the reason and then make a judgment).  \\
According to the command of the teammate, {[}Analysis{]}: The current command needs to be kept by `{[}XX term{]}'.   \\
If you see phrases like {[}Context{]} in answer template, replace the entire phrase according to the meaning of the Context, do not repeat the content; make analogy expansion for `...'; keep `:'; absolutely do not modify others in template. \\ \hline
\end{tabular}
\caption{Chain of thought prompt for GPT4.}
\label{tab_gpt4_prompt}
\end{table}

\begin{table}[h]
\centering
\begin{tabular}{|l|l|l|}
\hline
\multirow{6}{*}{prompt} & \begin{tabular}[c]{@{}l@{}}system\\ background\\ prompt\end{tabular} & \begin{tabular}[c]{@{}l@{}}We have an agent and a player working together as a teammate in a \\ PUBG game. We hope you can help the agent plan how the agent's\\ game state should change, so as to complete the player's command and\\ help the player win the game.\end{tabular}                                                                                                                                                                                                                                                                                                                                                                                                                                                                                                                                                                                                                                                                                                                                                                                                                                                                                                                                                                                                                                                                                                                                     \\ \cline{2-3} 
                        & \begin{tabular}[c]{@{}l@{}}teammate\\ state\\ prompt\end{tabular}    & \begin{tabular}[c]{@{}l@{}}The state of the agent's teammates can be described as follows:\{`\\ Length of distance moved': `No movement', `Average velocity': `Slow',\\ `Horizontal direction of movement': `Southeast', `Horizontal \\ direction of view': `South', `Pitch direction of view': `Medium', \\ `Health level': `Empty', `Whether to restore health': `False', \\ `Whether the health is damaged': `False', `Whether rescued teammate':\\  `False', `Whether prone position': `False', `Whether crouch \\ position': `False', `Whether have a gun': `True', `Whether hold a \\ gun': `False', `Whether have bullets': `True', `Whether have medical\\ kits': `True', `Whether be knocked down': `False', `Damage to enemy': \\ `Zero', `Whether knock down enemy': `False', `Whether seen enemy': \\ `True', `Number of enemies have ever seen': 5, `Whether seen by \\ enemy': `True', `Distance with nearest enemy': `Nearby', `Whether \\ closer with nearest enemy': `False', `ID of teammate player': 2\}\end{tabular}                                                                                                                                                                                                                                                                                                                                                              \\ \cline{2-3} 
                        & \begin{tabular}[c]{@{}l@{}}enemy\\ state\\ prompt\end{tabular}       & \begin{tabular}[c]{@{}l@{}}The state of the enemy can be described as follows:\{`Horizontal \\ direction of movement of enemy': `Southwest', `Velocity of enemy':\\ `Slow', `Enemy's position relative to agent': `West'\}\end{tabular}                                                                                                                                                                                                                                                                                                                                                                                                                                                                                                                                                                                                                                                                                                                                                                                                                                                                                                                                                                                                                                                                                                                               \\ \cline{2-3} 
                        & \begin{tabular}[c]{@{}l@{}}self\\ state\\ prompt\end{tabular}        & \begin{tabular}[c]{@{}l@{}}The state of the agent can be described as follows:\{`Damage to \\ enemy': `Zero', `Whether knock down enemy': `False', `Whether kill \\ enemy': `False', `Whether seen enemy': `True', `Whether seen by \\ enemy': `True', `Number of enemies have ever seen': 3, `Length of \\ distance moved': `Short', `Average velocity': `Fast', `Horizontal \\ direction of movement': `West', `Horizontal direction of view': \\ `NorthEast', `Pitch direction of view': `Medium', `Health level': \\ `Full', `Whether to restore health': `False', `Whether the health is\\  damaged': `False', `Whether rescued teammate': `False', `Whether be\\  knocked down': `False', `Whether prone position': `False', `Whether\\  have a gun': `True', `Whether have bullets': `True', `Whether have \\ medical kits': `True', `Distance with nearest enemy': `Nearby', \\ `Whether closer with nearest enemy': `True', `Whether crouch \\ position': `False', `Whether hold a gun': `True', `Whether seen by \\ teammate': `True', `Length of distance from agent to teammate': \\ `Touch', `Teammate's position relative to agent': `Southwest', \\ `Whether follow with the views of teammate': `False', `Whether \\ target the same enemy as teammate': `False', `Whether follow with \\ the movement direction of teammate': `False'\}\end{tabular} \\ \cline{2-3} 
                        & \begin{tabular}[c]{@{}l@{}}command\\ prompt\end{tabular}             & \begin{tabular}[c]{@{}l@{}}Teammate player gives you a verbal command based on the current\\ game states:`You should lie in wait.'\end{tabular}                                                                                                                                                                                                                                                                                                                                                                                                                                                                                                                                                                                                                                                                                                                                                                                                                                                                                                                                                                                                                                                                                                                                                                                                                       \\ \cline{2-3} 
                        & \begin{tabular}[c]{@{}l@{}}question\\ prompt\end{tabular}            & \begin{tabular}[c]{@{}l@{}}In order to complete the command `You should lie in wait.', how\\ the agent's game state should change?\end{tabular}                                                                                                                                                                                                                                                                                                                                                                                                                                                                                                                                                                                                                                                                                                                                                                                                                                                                                                                                                                                                                                                                                                                                                                                                                                        \\ \hline
response                & \begin{tabular}[c]{@{}l@{}}goal\\ meta-state\\ prompt\end{tabular}   & \begin{tabular}[c]{@{}l@{}}Whether prone position:True\\ Average velocity:Static\\ Length of distance moved:No movement\\ Whether hold a gun:True\end{tabular}                                                                                                                                                                                                                                                                                                                                                                                                                                                                                                                                                                                                                                                                                                                                                                                                                                                                                                                                                                                                                                                                                                                                                                                     \\ \hline
\end{tabular}
\caption{Example of prompt and response.}
\label{tab_prompt}
\end{table}



\end{document}